\documentclass[journal]{IEEEtran}
\ifCLASSINFOpdf
\else
\fi
\usepackage{times}
\usepackage{epsfig}
\usepackage{graphicx}
\usepackage{amsmath}
\usepackage{amssymb}
\usepackage{bbm}
\usepackage{color}
\usepackage{float}
\usepackage{booktabs}
\usepackage{multirow}% http://ctan.org/pkg/multirow
\usepackage{hhline}% http://ctan.org/pkg/hhline
\usepackage{ltablex}% <-- added
\usepackage{algorithm,algorithmic}

\newcommand{\RNum}[1]{\uppercase\expandafter{\romannumeral #1\relax}}
\newcommand{\etal}[1]{\textit{et al.} \cite{#1}}

\DeclareMathOperator{\Flow}{Flow}
\DeclareMathOperator{\warp}{warp}
\DeclareMathOperator{\Aug}{Aug}
\DeclareMathOperator{\Enc}{Enc}
\DeclareMathOperator{\Deconv}{Deconv}
\DeclareMathOperator{\Conv}{Conv}
\DeclareMathOperator{\softmax}{softmax}
\DeclareMathOperator{\GarmentNet}{GarmentNet}
\DeclareMathOperator{\SynthesisNet}{SynthesisNet}

\newcommand{\norm}[1]{{\left|\left|#1\right|\right|}}

\begin{document}
%
% paper title
% Titles are generally capitalized except for words such as a, an, and, as,
% at, but, by, for, in, nor, of, on, or, the, to and up, which are usually
% not capitalized unless they are the first or last word of the title.
% Linebreaks \\ can be used within to get better formatting as desired.
% Do not put math or special symbols in the title.
\title{Unsupervised Pose Flow Learning for Pose Guided Synthesis}
%
%
% author names and IEEE memberships
% note positions of commas and nonbreaking spaces ( ~ ) LaTeX will not break
% a structure at a ~ so this keeps an author's name from being broken across
% two lines.
% use \thanks{} to gain access to the first footnote area
% a separate \thanks must be used for each paragraph as LaTeX2e's \thanks
% was not built to handle multiple paragraphs
%

\author{Haitian Zheng,~\IEEEmembership{Student Member,}
        Lele Chen,~\IEEEmembership{Student Member,}
        Chenliang Xu,~\IEEEmembership{Member,}
        and~Jiebo~Luo,~\IEEEmembership{Fellow, IEEE}% <-this % stops a space}
    }

%\thanks{H. Zheng, L. Chen and J.Luo are with the Department of Computer Science, University of Rochester, Rochester, NY, 14627 USA (e-mail: hzheng15@ur.rochester.edu; lchen63@cs.rochester.edu; jluo@cs.rochester.edu).}
%\thanks{Corresponding author: Jiebo Luo.}

% note the % following the last \IEEEmembership and also \thanks - 
% these prevent an unwanted space from occurring between the last author name
% and the end of the author line. i.e., if you had this:
% 
% \author{....lastname \thanks{...} \thanks{...} }
%                     ^------------^------------^----Do not want these spaces!
%
% a space would be appended to the last name and could cause every name on that
% line to be shifted left slightly. This is one of those "LaTeX things". For
% instance, "\textbf{A} \textbf{B}" will typeset as "A B" not "AB". To get
% "AB" then you have to do: "\textbf{A}\textbf{B}"
% \thanks is no different in this regard, so shield the last } of each \thanks
% that ends a line with a % and do not let a space in before the next \thanks.
% Spaces after \IEEEmembership other than the last one are OK (and needed) as
% you are supposed to have spaces between the names. For what it is worth,
% this is a minor point as most people would not even notice if the said evil
% space somehow managed to creep in.

% The paper headers
%\markboth{IEEE Transactions on Image Processing, under review}%
\markboth{~}%
{Shell \MakeLowercase{\textit{et al.}}: Bare Demo of IEEEtran.cls for IEEE Journals}

% The only time the second header will appear is for the odd numbered pages
% after the title page when using the twoside option.
% 
% *** Note that you probably will NOT want to include the author's ***
% *** name in the headers of peer review papers.                   ***
% You can use \ifCLASSOPTIONpeerreview for conditional compilation here if
% you desire.

% If you want to put a publisher's ID mark on the page you can do it like
% this:
%\IEEEpubid{0000--0000/00\$00.00~\copyright~2015 IEEE}
% Remember, if you use this you must call \IEEEpubidadjcol in the second
% column for its text to clear the IEEEpubid mark.

% use for special paper notices
%\IEEEspecialpapernotice{(Invited Paper)}

% make the title area
\maketitle

% As a general rule, do not put math, special symbols or citations
% in the abstract or keywords.
\begin{abstract}
	Pose guided synthesis aims to generate a new image in an arbitrary target pose while preserving the appearance details from the source image. Existing approaches rely on either hard-coded spatial transformations or 3D body modeling. They often overlook complex non-rigid pose deformation or unmatched occluded regions, thus fail to effectively preserve appearance information. In this paper, we propose an unsupervised pose flow learning scheme that learns to transfer the appearance details from the source image. Based on such learned pose flow, we proposed GarmentNet and SynthesisNet, both of which use multi-scale feature-domain alignment for coarse-to-fine synthesis. Experiments on the DeepFashion, MVC dataset and additional real-world datasets demonstrate that our approach compares favorably with the state-of-the-art methods and generalizes to unseen poses and clothing styles.
\end{abstract}

% Note that keywords are not normally used for peerreview papers.

% For peer review papers, you can put extra information on the cover
% page as needed:
% \ifCLASSOPTIONpeerreview
% \begin{center} \bfseries EDICS Category: 3-BBND \end{center}
% \fi
%
% For peerreview papers, this IEEEtran command inserts a page break and
% creates the second title. It will be ignored for other modes.
\IEEEpeerreviewmaketitle

\section{Introduction}
\IEEEPARstart{P}{ose} guided synthesis aims to generate a realistic person image that preserves the appearance details of the source image given an arbitrary target pose. As a central task in virtual reality \cite{flycap}, online garment retail \cite{viton}, and game character rendering, realistic pose guided synthesis will have a crucial impact on numerous applications.

Despite the recent successes of conditional image synthesis \cite{pix2pix,pix2pixHD}, pose guided synthesis still faces many unsolved challenges. Among them, the main challenge is the complex, part-independent pose deformation, with garment, from the source pose to an arbitrary target pose. 
As a result, models \cite{pg2,vunet,viton,swapnet} built on the plain U-Net \cite{unet} network structure often fail to generate precise details or textures due to the lack of a robust spatial alignment component.

Recently, several approaches \cite{deform_gan,densepose_transfer,m2e,softgated} have been proposed to address spatial alignment. Specifically, Siarohin \etal{deform_gan} apply deformable skip connections for spatial alignment. However, the oversimplified affine transformation on the predefined rectangles does not necessarily capture the non-rigid deformation. Different from Siarohin \textit{et al.}, Neverova \etal{densepose_transfer} and Wu \etal{m2e} resort to a pretrained pose estimator, DensePose \cite{densepose}, to perform non-rigid alignment on 3D-model. Since such model-level alignment is not capable of handling occluded regions caused by drastic pose changes, inpainting is then applied to fill the occluded region. Nonetheless, the results are usually blurry in occluded regions.
% , as the details provided by source images are not properly utilized. 
\begin{figure}[t]
\centering
\includegraphics[width=1.0\linewidth]{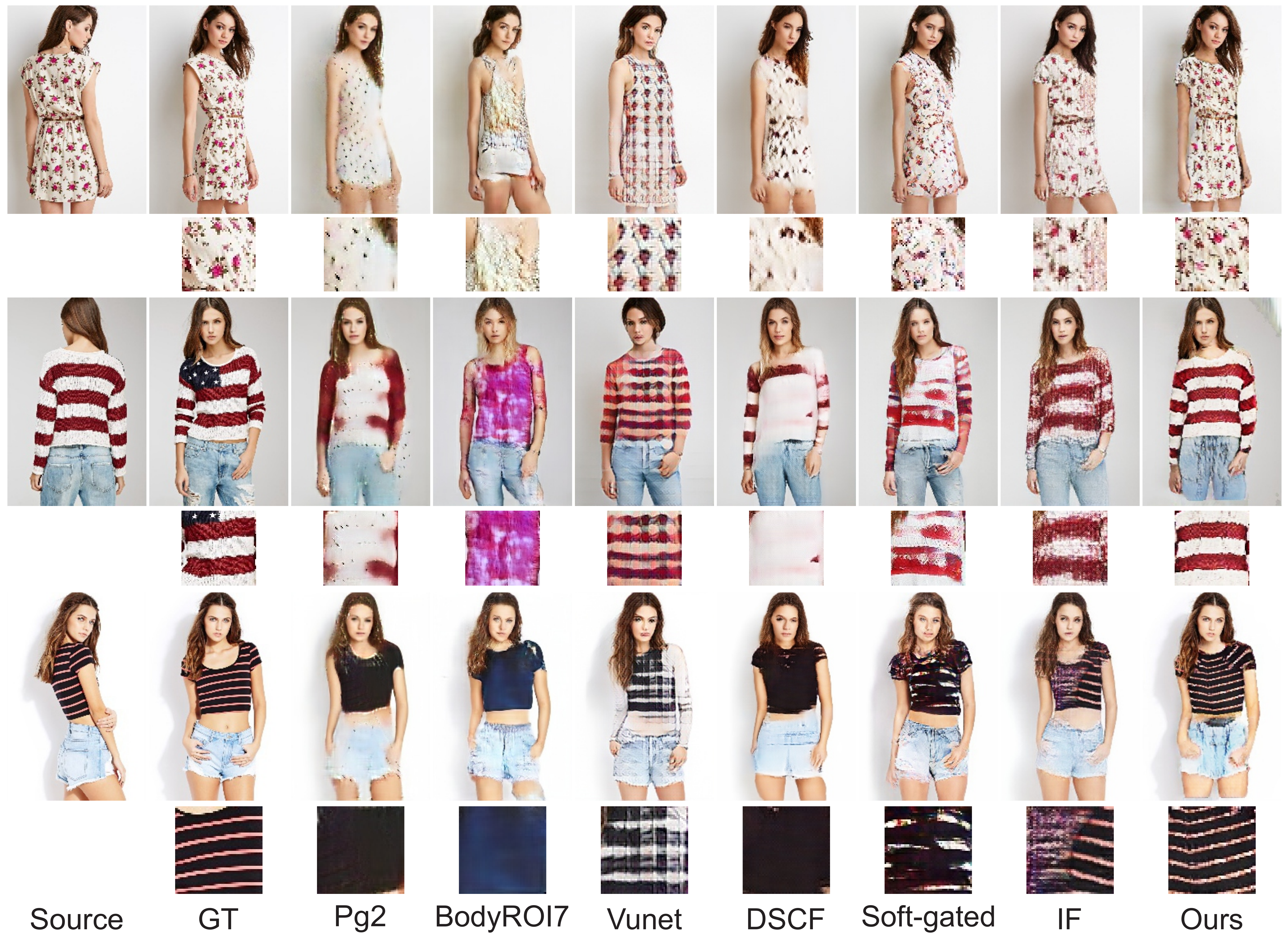}
\caption{Images generated by different methods. The first column contains source images while the second column contains ground truth images with target poses. We compare our results (last column) with the state-of-the-art methods (rows 3-7). The odd rows display the entire images and the even rows display the corresponding texture details. In comparison, our method clearly  produces the most visually plausible and pleasing effects.}
\label{fig:teaser2}
% \vspace{-3mm}
\end{figure}

\begin{figure*}[t]
	\centering
	\includegraphics[width=1.0\textwidth]{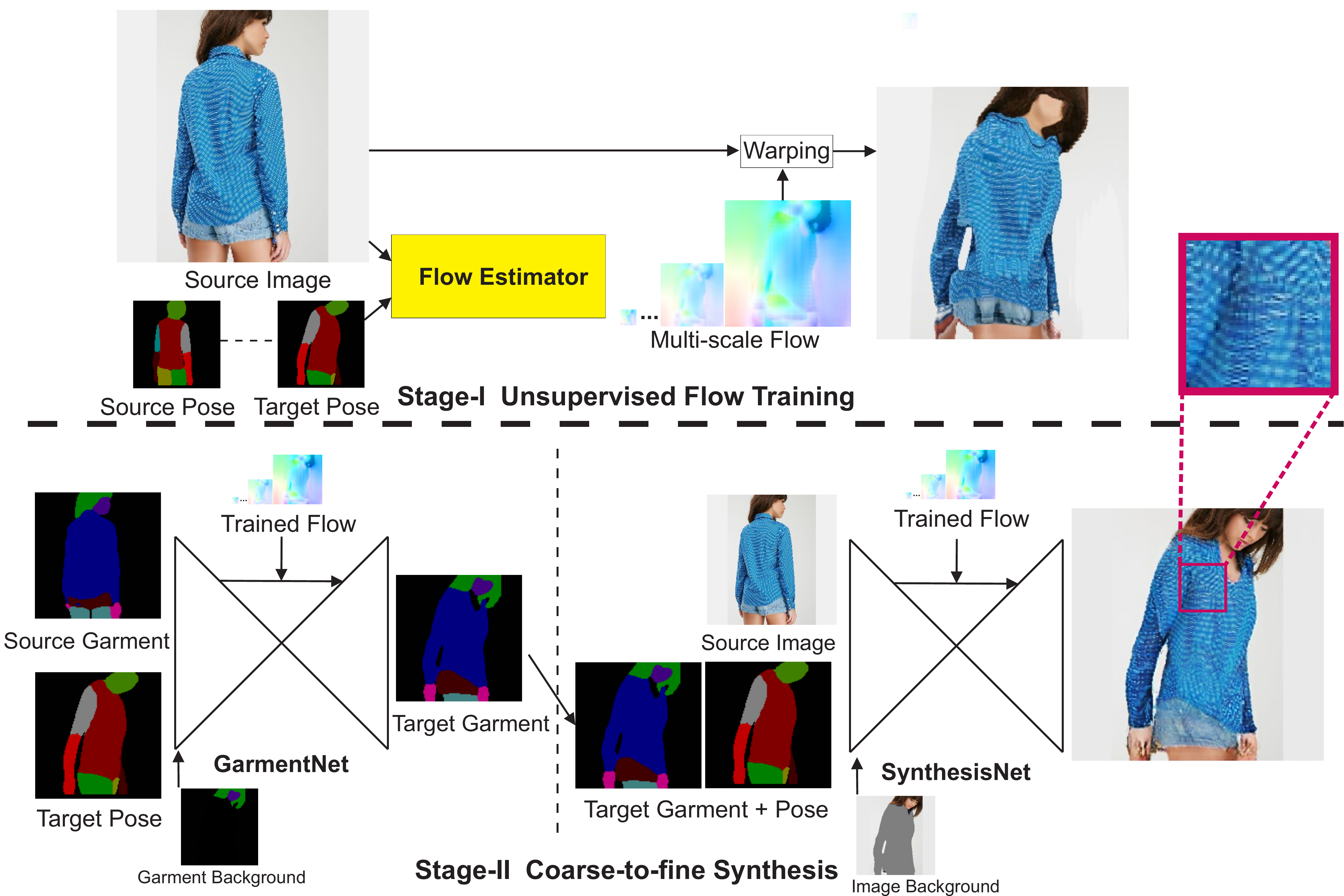}
	\caption{Our two-stage framework for pose-guided person image synthesis. In stage-\RNum{1}, a flow estimator is trained using our proposed texture-preserving objective. In stage-\RNum{2}, GarmentNet and SynthesisNet use the trained flow estimator to sequentially estimate garment parsing and image output, following a course-to-fine pipeline.}
%		The framework of the proposed Garment2PoseNet. 
	%More adaptive comparing to affine transformation, and handling occlusion: A key observation we made is that, even in the occluded region, the crucial texture information for rendering can be implicitly preserved from other body parts. For instance, by looking at the back of garment, we are able to guess the appearance of the front. A learned flow is therefore adaptive enough to preserve texture for occluded region from other regions.\lele{reedit this figure}
	\label{fig:pipeline}
\end{figure*}

A later work \cite{softgated} relies on the combination of affine transformation and thin-plate splines (TPS) transformation to perform spatial alignment. However, the TPS transformation is inflexible to model the highly non-rigid human pose deformation. In addition, their matching module is trained on simplified synthetic transformations \cite{geo}. Therefore, the human pose deformation is not properly handled. Most recently, Li \etal{intrinsic_flow} use the 3D human model \cite{SMPL} to generate human pose flow ground-truth for training a flow estimator. 
However, similar to other 3D-modeling approaches \cite{densepose_transfer,m2e}, the issue of large occluded regions is not well addressed due to the lack of correspondence. Moreover, the 3D human modeling is computationally expensive, and it is not always precise on loose clothes, as 3D human modeling focus on body reconstruction rather than the clothes surface reconstruction.

In this paper, we present i) a novel unsupervised pose flow learning scheme (Stage-\RNum{1}) to tackle the pose guided transfer task. Next, we propose ii) a coarse-to-fine garment-to-image synthesis pipeline (Stage-\RNum{2}) using feature domain alignment based on the learned flow. Without using affine or TPS transformation \cite{deform_gan,softgated} or resorting to explicit 3D human modeling \cite{densepose_transfer,m2e,intrinsic_flow} to extract correspondence, our method utilizes learned pose flow to capture the complex pose deformation. To address the issue of occlusion caused by drastic pose changes, we propose an unsupervised pose flow learning scheme that learns to transfers appearance to occluded regions. In contrast to \cite{intrinsic_flow}, our approach avoids the computationally inefficient flow ground-truth generation step.

To enable such an unsupervised pose flow training scheme, we propose in Stage-\RNum{1} a novel texture preserving objective to improve the quality of the learned flow, which is shown to be crucial for the pose-guided synthesis task. We also propose augmentation-based self-supervision to stabilize the flow training. Based on the learned pose flow, we proposed in Stage-\RNum{2} a coarse-to-fine garment-to-image synthesis pipeline using our proposed GarmentNet and SynthesisNet. GarmentNet and SynthesisNet share a unified network structure, which utilizes the learned pose flow for multi-scale feature domain warping. Furthermore, we propose a novel gated multiplicative attention module for misalignment-aware synthesis. 

Finally, to synthesize more realistic images,  we design masking layers in GarmentNet and SynthesisNet to preserve the target image background and person identity for realistic synthesis. Furthermore, we use DensePose parsing~\cite{densepose} instead of person keypoints as pose inputs. DensePose parsing contains body segmentation and mesh coordinates, which provide richer information for realistic pose-guided synthesis.

\vspace{10pt}
Our main contributions are three-fold:
\begin{itemize}
    \item We propose an unsupervised pose flow learning scheme for pose-guided synthesis. Our scheme adaptively learns to transfer appearance from target images. To enable such a learning scheme, a novel texture preserving objective and an augmentation-based self-supervision strategy are proposed, which improve the quality of the transferred appearance. 
    \item We propose a coarse-to-fine synthesis pipeline based on GarmentNet and SynthesisNet. GarmentNet and SynthesisNet are based on the learned pose flow for multi-scale feature domain alignment. Furthermore, a novel gated multiplicative attention module is proposed to address the misalignment issue.
    \item To facilitate more realistic image synthesis, we design masking layers that preserve target identities and background information. Furthermore, we use DensePose parsing as pose representation, which provides richer pose details for pose-guided synthesis.
\end{itemize}
% \begin{itemize}
%     \item We propose a flow-based alignment scheme for pose guided synthesis to handle complex pose deformation and texture transfer for occluded region synthesis. This scheme is unsupervised which does not require the time consuming flow annotation step.
%     \item We propose a coarse-to-fine synthesis scheme consists of  GarmentNet and SynthesisNet for garment parsing generation and pose-guided image synthesis. GarmentNet and SynthesisNet are based on a unified network structure for flow-based alignment.
%     \item Our approach achieves both qualitatively and quantitatively better results in comparison to the state-of-the-art methods on the DeepFashion dataset and the Multi-view Clothing dataset. Further evaluations show that our approach can generalize to real-world scenario.
% \end{itemize}

The remainder of the paper is organized as follows. Sec.~\ref{sec:related} introduces related work on (pose guided) image synthesis and optical flow learning. The proposed approach is detailed in Sec.~\ref{sec:approach}. Experiments are described in Sec.~\ref{sec:experiment}. Sec.~\ref{sec:conlusion} concludes the paper.

\section{Related Work}
\label{sec:related}
\subsection{Image synthesis}
Generative Adversarial Network (GAN)~\cite{gan} has been widely used for image synthesis tasks. Conditional GAN~\cite{cgan} aims to synthesize an image from an given conditional input content. Based on conditional GAN, Isola \textit{et al.}~propose Pix2Pix~\cite{pix2pix} for image style transfer tasks. Later on, many techniques have been proposed to improve both the synthesis quality and resolution of the generated images. Specifically, Johnson \textit{et al.}~\cite{perceptual_loss} use feature-level distance on the VGG network~\cite{vgg} to measure the perceptual similarities. The Gram matrix loss~\cite{gram} is proposed by Gatys \textit{et al.} for texture synthesis. To improve the image synthesis resolution, Zhang \textit{et al.}~\cite{stackgan} propose a two-stage network for generating images from coarse to fine scales. PatchGAN discriminator~\cite{patchgan} is used by Li \textit{et al.}~to penalize unrealistic patches. Wang \textit{et al.}~\cite{pix2pixHD} and Chen \textit{et al.}~\cite{crn} propose new generator structures for realistic image synthesis. In addition, techniques such as Wasserstein distance~\cite{wgan} and Spectral Normalization~\cite{sn} are proposed to stabilize GAN training. Those approaches have improved the synthesized image quality. However, these approaches are limited to spatial deformation as their networks are built on local convolution. In this work, we present a flow-based approach to address the spatial alignment problem in pose-guided synthesis.

\subsection{Pose Guide Synthesis}
 Ma \textit{et al.}~\cite{pg2} use  the source image and target pose landmarks as the conditional input and the UNet~\cite{unet} structure for pose guided synthesis. Later, Siarohin \textit{et al.}~\cite{deform_gan} utilize skip connections with hard-coded part-level affine transformation to transform feature maps for new pose image synthesis. Dong \textit{et al.}~\cite{softgated} use the thin-plate spline (TPS) transform trained on synthetic transformations~\cite{geo} to warp the source domain content. Additionally, Han \textit{et al.}~\cite{viton} and Wang \textit{et al.}~\cite{CP-VTON} use the TPS transformer for virtual try-on. To handle pose deformations, Neverova \textit{et al.}~\cite{densepose_transfer} use DensePose~\cite{densepose} to transfer appearance patterns and utilize in-painting to fill occluded regions. In addition, pose guided synthesis is formulated as a  pose-appearance disentanglement problem. Specifically, Esser \textit{et al.}~\cite{vunet} use variational autoencoder~\cite{vae} to capture the latent space of pose and appearance for appearance manipulation under given poses. Ma \textit{et al.}~\cite{disentangle_gan} learn disentangled pose-appearance representation using a multi-branch encoding and decoding scheme. However, the plain UNet structure~\cite{pg2,vunet}, predefined transformation~\cite{deform_gan,densepose_transfer} or TPS transformer~\cite{softgated,CP-VTON} are insufficient for handling the complex human pose deformation and occlusion caused by drastic pose changes. Recently, Li \textit{et al.}~\cite{intrinsic_flow} uses 3D human model~\cite{SMPL} to correspondence annotation, then fit a flow estimator to speed up inference. However, generating the correspondence supervision is computationally exhausted. Furthermore the ground-truth correspondence cannot effectively transfer appearance to occluded regions. In contrast, our unsupervised flow-training scheme learns to transfer appearance under complex pose deformation and occlusion without using explicit correspondence annotation.

\subsection{Unsupervised Optical Flow Learning}
Recently, several approaches have been proposed to learn optical flow in the absence of the ground-truth annotation. Specifically, Jason \textit{et al.}~\cite{back-to-basic} optimize a predictive model using a combination of photometric loss and smoothness. Meister \textit{et al.}~\cite{unflow} utilize left-right consistency to filter out occluded regions.  Wang \textit{et al.}~\cite{occlusion-aware} further propose an occlusion-aware objective function for unsupervised flow learning. Different from these works, we focus on learning a flow that better preserves the appearance information. Furthermore, our optical flow is estimated using only the source image and pose information.

\section{Approach}
\label{sec:approach}
In this section, we present an unsupervised flow-based approach to the pose-guided synthesis task. To this end, we adopt a two-stage pipeline, as illustrated in Fig.~\ref{fig:pipeline}. In Stage-\RNum{1}, a flow estimator is unsupervisedly trained using our proposed texture-preserving objective. In Stage-\RNum{2}, we present GarmentNet and SynthesisNet to sequentially generate garment parsing and image output, using the flow obtained from the previous stage. 

In Sec.~\ref{subsection:notation}, we first define the notation that are required by our model.
In Sec.~\ref{subsection:method-stage1}, we propose our unsupervised texture-preserving objective and other details for training flow estimator for pose-guided alignment.
In Sec.~\ref{subsection:method-stage2}, we propose GarmentNet and SynthesisNet to respectively estimate garment parsing and image output. 

\subsection{Notations}
\label{subsection:notation}
\begin{figure}[]
	\centering
	\includegraphics[width=0.65\linewidth]{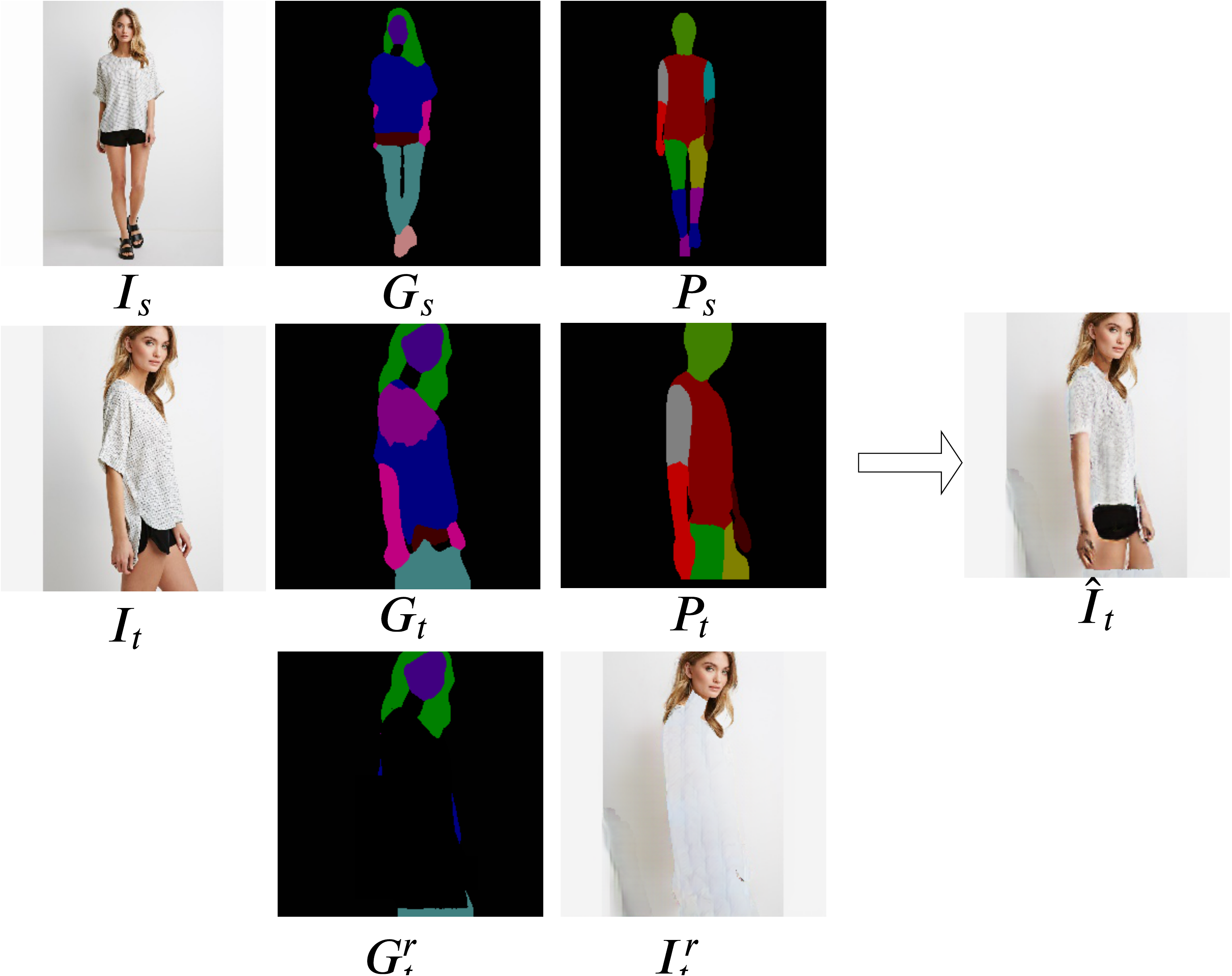}
	\caption{Notation illustrations for the required data for training and testing. We use subscripts $s$ and $t$ to represent source and target domains, respectively. The notions of $I$, $G$ and $P$ represent images, garment parsing and pose representation, respectively.  $(I^{r}_t, G^{r}_t)$ denote image residue and garment residue from the target person. The output of our approach is denoted by $\hat{I}_t$. Please refer to Sec.~\ref{subsection:notation} for more details.}.
	\label{fig:data}
\end{figure}

Given a pair of images $I_s$ and $I_t$ from the source and target domains respectively, pose-guided synthesis aims to generate a image $\hat{I}_t$ that preserves the appearance of $I_s$ and the pose of $I_t$. To this end, we respectively generate \textit{pose representation} $P_s, P_t$ and \textit{garment parsing} $G_s, G_t$ from $I_s$ and $I_t$, to capture useful information from the source and target domains. In addition, we extract image residue $I^r_t$ from $I_t$ and garment residues $G^{r}_t$ from garment $G_t$, in the hope to capture target identity (i.e., face, hair, and background regions). Fig.~\ref{fig:data} illustrates $(P_s, P_t)$, $(G_s, G_t)$, $(I_s, I_t)$ and residues $(I^{r}_t, G^{r}_t)$. In fact, $P_t, G_t$ and $ I_t$ form an hierarchical structure that gradually provide richer information of the target person. We leverage this hierarchical structure in Sec.~\ref{subsection:method-stage2} to design our coarse-to-fine synthesis pipeline. We note that during training, $I_s$ and $I_t$ are from the same outfit of the same person. In testing phase, however, $I_s$ and $I_t$ can be  arbitrary person with arbitrary outfits.

To be more specific, the pose representations $P_s$ and $P_t$ are the concatenation of the one-hot pose parsing and the mesh coordinate map from Densepose~\cite{densepose}. Likewise, the garment representations $G_s$ and $G_t$ are the one-hot garment parsing generated using the method by Gong \etal{LIP}. The image residue $r^{i}_t$ are generated by first removing person region from $I_t$ then perform inpainting \cite{}. Then, hair and face regions are appended on the inpainted results \footnote{We use the garment parsing $G_t$ to generate the regions of human body, hair and face.}. Finally, garment residue $r^{g}_t$ are generated by setting values of one-hot parsing $G_t$ to $0$ for background, face and hair channels.

Although our approach can adapt key-point heat maps as an alternative human pose representation, we argue that sparse key-points do not provide sufficient pose information for accurate person image generation. By contrast, DensePose parsing and mesh coordinates provide dense, pseudo-3D information, which is informative to represent pose detail.

%As illustrated in Figure \ref{fig:pipeline}, a two-stage approach is proposed to transfer the appearance details from the source pose to the target pose for pose guided synthesis.
%In Stage-\RNum{1}, a flow estimator is trained to capture pose deformations using the proposed unsupervised appearance preserving objective function. In Stage-\RNum{2}, the GarmentNet and SynthesisNet are designed to sequentially generate target garment parsing and final output in a coarse-to-fine manner. The GarmentNet and SynthesisNet share a unified network structure, which relies on the trained flow estimator for multi-scale feature domain alignment (as elaborated in Section \ref{subsection:method-stage2}). In addition, a multiplicative attention fusion module is  proposed to handle feature misalignment.

%In Section \ref{subsection:method-stage1}, we describe the flow estimator and the unsupervised objective for training the appearance preserving flow. In Section \ref{subsection:method-stage2}, the GarmentNet and SynthesisNet as well as the  multiplicative attention fusion module are elaborated.

\begin{figure}[t]
	\centering
	\includegraphics[width=1.0\linewidth]{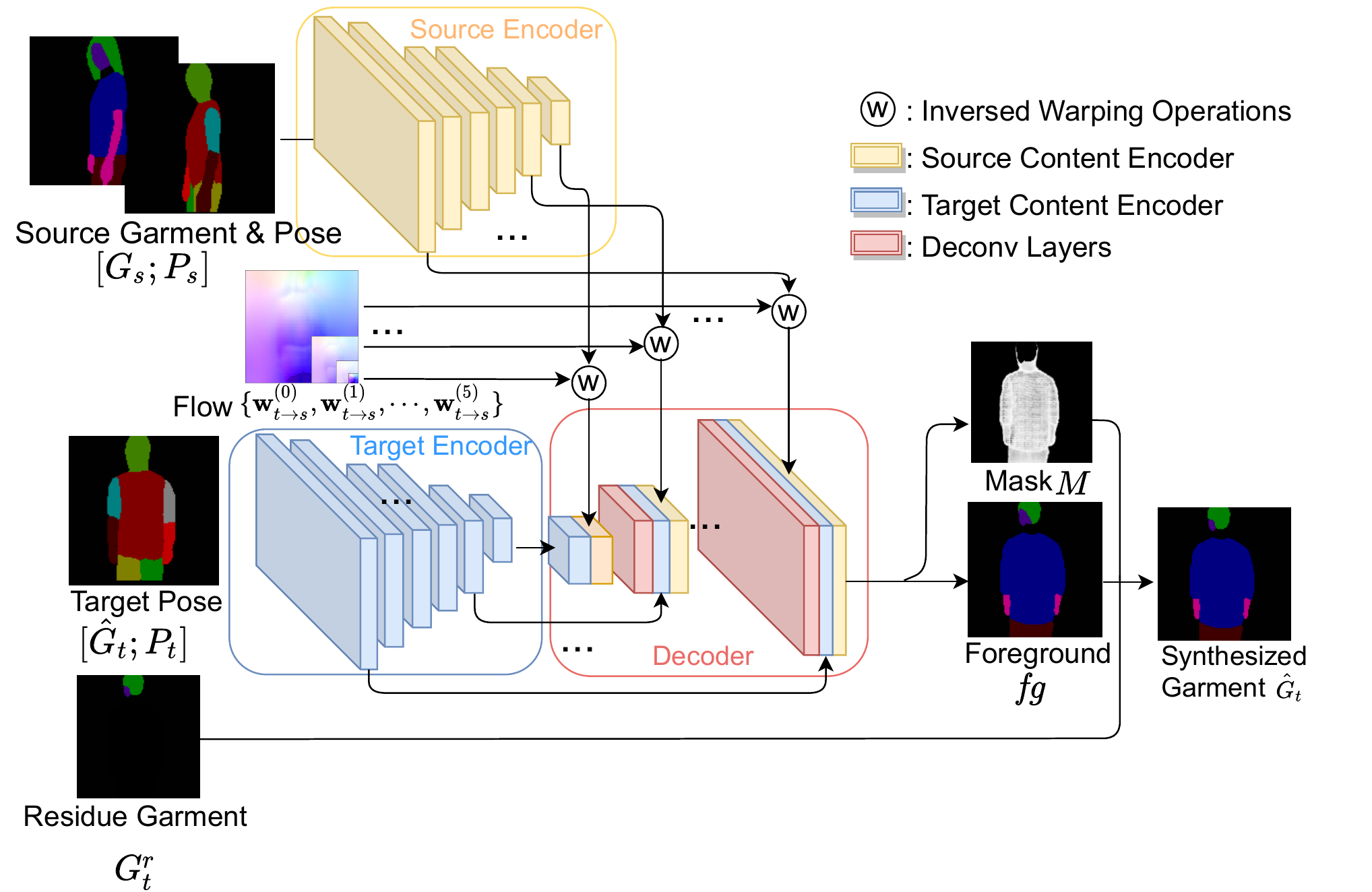}
	\caption{The network structure of GarmentNet. Given the generated flow from Stage-\RNum{1}, GarmentNet encodes information from the source and target domains using a Source Domain Encoder (yellow) and a Target Domain Encoder (blue), respectively. After warping-based alignment, the source domain features are aggregated with the target domain features at multiple scales by our Decoder (red). Finally, the generated foreground is alpha-blended with the residue garment to synthesize garment parsing. In testing stage, the source and target image are from different persons.}
	\label{fig:GarmentNet}
\end{figure}

\subsection{Stage-\RNum{1}: Unsupervised Texture Preserving Flow}
\label{subsection:method-stage1}
With the extracted pose representations $P_s$ and $P_t$, we present an unsupervised flow training scheme to generate adaptive, texture-preserving alignment without resorting to the computationally inefficient SMPL model \cite{SMPL} or oversimplified affine~\cite{deform_gan} or TPS transformation~\cite{CP-VTON,softgated}. 
%Instead of using affine~\cite{deform_gan} or TPS transformation~\cite{CP-VTON,softgated} or resorting to the computationally inefficient SMPL model\cite{SMPL} to annotate correspondences, we rely on an unsupervisedly trained flow estimator for adaptive, texture-preserving alignment. 
%To enable such training
% using only image pairs as annotation. To enable such 
%is unsupervised, and it does not require the computationally expensive correspondence annotation.

As shown in Fig.~\ref{fig:pipeline}, our flow estimator takes the source image, pose and target pose as inputs to generate multi-scale flow-fields to indicate the pose deformation. Formally, let ${\Flow}(\cdot, \cdot)$ denote our flow estimator, which takes $[I_s;P_s]$ and $P_t$ from source and target domains as inputs and outputs flow fields at multiple scales:
\begin{equation}
	\{\mathbf{w}^{(0)}_{t\rightarrow s}, \mathbf{w}^{(1)}_{t\rightarrow s}, \cdots, \mathbf{w}^{(5)}_{t\rightarrow s}\} = {\Flow}([I_s;P_s],P_t).
\label{equ:flow}
\end{equation} 
where notation $\mathbf{w}^{(l)}_{t\rightarrow s}$ denotes flow field from the target image to the source images at scale $l\in\{0,\cdots,5\}$.

We employ FlowNetS~\cite{flownet} as the baseline structure to implement ${\Flow}([I_s;P_s],P_t)$. Note that, unlike a normal flow estimator, ${\Flow}(\cdot,\cdot)$ leverages pose information for flow estimation. Meanwhile, we have also modified FlowNetS to improve the flow-field definition and to reduce memory usage. Please refer to Appendix~\ref{appendix:flownet} for more details.

Unsupervised flow training on natural images has been explored in several recent works. These approaches mainly rely on the photometric loss~\cite{back-to-basic}
\begin{equation}
\mathcal{L}_{p}(I_s,I_t,\mathbf{w}^{(0)}_{t\rightarrow s}) = \norm{\rho\left(I_t- \warp(I_s; \mathbf{w}^{(0)}_{t\rightarrow s})\right)}_1
\end{equation}
to measure the difference between the target image and the inversely warped source image using the predicted flow. Here, $\warp(\cdot;\cdot)$ denotes the inverse warping operation \cite{STN} and $\rho(x)={(x^2+\epsilon^2)}^\alpha$ is a robust loss function~\cite{charbonnier}. Furthermore, total variation-based (TV) spatial smoothness loss is also utilized to regularize the flow prediction \cite{ren2017unsupervised}:
\begin{equation}
\mathcal{L}_{TV}(\mathbf{w}^{(l)}_{t\rightarrow s}) =  {\norm{\frac{\partial}{\partial x}\mathbf{w}^{(l)}_{t\rightarrow s}}_1} + {\norm{ \frac{\partial}{\partial y}\mathbf{w}^{(l)}_{t\rightarrow s}}_1}.
\end{equation}
Due to the complexity of person images and the large displacement from source pose to target pose, the warping-based photometric term is highly non-convex. As as result, the gradient descendent training with the naive photometric loss and spatial smoothness loss will lead to difficulty in convergence. To solve this issue, we use multi-scale strategy, where photometric losses and spatial smoothness losses summed at multiple scales $l \in \{0,\dotsi ,5\}$. 

In our experiment, we found that the multi-scale training will still suffer from damaged local textures for the warped images $\warp(I_s; \mathbf{w}^{(0)}_{t\rightarrow s})$, and the learned flow fails to transfer realistic details from source images (see Fig.~\ref{fig:compare_flow} for details). We attribute this deficiency to the poor ability of $\mathcal{L}_{p}$ and $\mathcal{L}_{TV}$ in preserving the high-frequency texture. In order to preserve realistic details and textures for better pose-guided synthesis, we propose a texture-preserving objective $\mathcal{L}^{(l)}_{texture}$ that enforces texture similarity between the $I_t$ and $\warp(I_s; \mathbf{w}^{(0)}_{t\rightarrow s})$ at scale $l$:
%%%%%%%%%%%%%%%%%%%%%
\begin{equation}
\begin{aligned}
\label{eq:struct}
&\mathcal{L}^{(l)}_{texture}(I_t, I_s, \mathbf{w}^{(0)}_{t\rightarrow s})\\
= & \norm{ \mathbf{G}\left(\mathbf{f}_{vgg}^{(l)}(I_t)\right)-\mathbf{G}\left(\mathbf{f}_{vgg}^{(l)}(\warp(I_s; \mathbf{w}^{(0)}_{t\rightarrow s}))\right) }_1,
\end{aligned}
\end{equation}
%%%%%%%%%%%%%%%%%%%%%
where $\mathbf{f}_{vgg}^{(l)}(\cdot)$ represents the $l$'th VGG~\cite{vgg} feature map from layer \{\texttt{relu1\_2}, \texttt{relu2\_2},  \texttt{relu3\_2}, \texttt{relu4\_2}, \texttt{relu4\_3}\} of the given input image, and $\mathbf{G}(\cdot)$ denotes the Gram matrix~\cite{gram} to capture the second-order statistic of the given feature map. Although the objective $\mathcal{L}^{(l)}_{texture}$ is widely used in style transfer tasks, we are the first to show that the texture loss is crucial for learning a reasonable flow estimator for pose-guided synthesis tasks (see Fig.~\ref{fig:compare_flow} for details).

Finally, we use a multi-scale version of the three losses, which are then weighted summed to compute the final loss. Let $I^{(l)}_s$ and $I^{(l)}_t$ denote the resized images of $I_s$ and $I_t$ at scale $l \in \{0,\dotsi ,5\}$, the overall objective is given by: 
%%%%%%%%%%%%%%%%%%%%%%
\begin{equation}
\begin{aligned}
\label{eq:stage1_loss}
\mathcal{L}_{Stage\RNum{1}}& \\
= \sum_{l=0}^{5} s_l (&\mathcal{L}_{p} (I^{(l)}_s,I^{(l)}_t,\mathbf{w}^{(l)}_{t\rightarrow s})  \\ 
& + \beta_l \mathcal{L}^{(l)}_{texture}(I_t, I_s, \mathbf{w}^{(0)}_{t\rightarrow s}) \\
& + \gamma_l \mathcal{L}_{TV}(\mathbf{w}^{(l)}_{t\rightarrow s})),
\end{aligned}
\end{equation}
%%%%%%%%%%%%%%%%%%%%%%
with $(s_0, s_1, s_2, s_3, s_4)$ $=(1, 1, 0.5, 0.25, 0.125)$, $(\beta_0, \beta_1, \beta_2,$ $ \beta_3, \beta_4)$ $=(0.002, 0.002, 0.002, 0.002, 0)$, $(\gamma_0, \gamma_1, $ $ \gamma_2, \gamma_3, \gamma_4)$ $=(0.1, 0.1, 0.1, 0.1, 0)$.

To further stabilize the training, an augmentation-based self-supervision is employed to regularize the learned flow. Specifically, let $\Aug(\cdot,\theta)$ denote an augmentation transformation based on cropping, affine transformation and flipping with a random control parameter $\theta$, the augmented source pose and source image are treated as target pose and target image, respectively. More precisely, we use the following update rules to transform the original data before one iteration of flow estimator training: 
\begin{equation}
\begin{aligned}
\epsilon &\sim U(0,1),\\
P_t &\gets \Aug(P_s,\theta),\text{if $\epsilon<0.25$}\\
I_t &\gets \Aug(I_s,\theta),\text{if $\epsilon<0.25$}.
\end{aligned}
\end{equation}
During training, $25\%$ percent of the training samples are first generated using such a synthetic random transformation to help the flow estimator to learn from simple transformations.

\subsection{Stage-\RNum{2}: Coarse-to-Fine Synthesis}
\label{subsection:method-stage2}
\begin{figure}[t]
	\centering
	\includegraphics[width=1.05\linewidth]{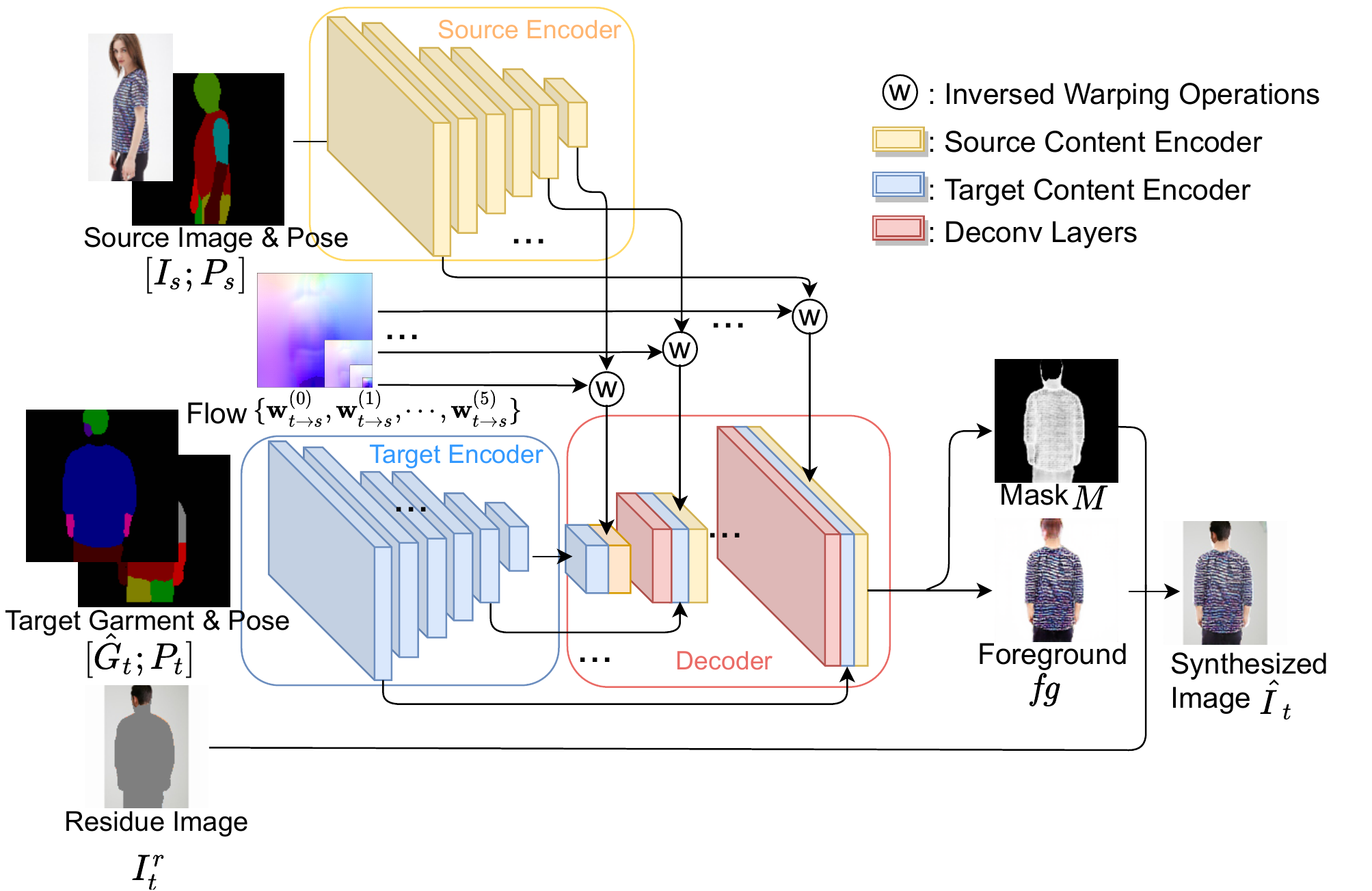}
	\caption{The network structure of SynthesisNet. Given the generated flow from Stage-\RNum{1} and the synthesized garment parsing, GarmentNet encodes information from the source and target domains using a Source Domain Encoder (yellow) and a Target Domain Encoder (blue), respectively. After warping-based alignment, the source domain features are aggregated with the target domain features at multiple scales by the Decoder (red). Finally, the generated foreground is alpha blended with the residue image to synthesize image output. In testing stage, the source and target image are from different persons.}
	\label{fig:SynthesisNet}
\end{figure}
Based on the learned flow estimator in Stage-\RNum{1}, we propose GarmentNet and SynthesisNet to sequentially synthesize garment parsing and image output following a coarse-to-fine pipeline (Fig. \ref{fig:pipeline} bottom). As illustrated in Fig.~\ref{fig:GarmentNet} and Fig.~\ref{fig:SynthesisNet}, GarmentNet and SynthesisNet share a unified network structure, which utilize the learned flow in stage-\RNum{1} for feature alignment. Afterwards, U-Net decoder serves to fuse information from both the source and target domains. On top of the decoder, an alpha blending layer is applied to preserve background information and to generate final outputs.

Formally, $\GarmentNet$ utilizes $[G_s, P_s]$ to encode source domain information, $P_t$ to encode target domain information, $\{\mathbf{w}^{(0)}_{t\rightarrow s}, \mathbf{w}^{(1)}_{t\rightarrow s}, \cdots, \mathbf{w}^{(5)}_{t\rightarrow s}\}$ from stage-\RNum{1} for alignment, and $G^{r}_t$ to keep the shape of target hair and face. The notation $[\cdot,\cdot]$ denotes channal-wise concatenation. The output target garment of $\GarmentNet$ is denoted by $\hat{G}_t$:
%%%%%%%%%%%%%%%%%%%%%%
\begin{equation}
\label{eq:clothnet}
\begin{aligned}
	\hat{G}_t=\GarmentNet(&[G_s, P_s], P_t,\\ &\{\mathbf{w}^{(0)}_{t\rightarrow s}, \mathbf{w}^{(1)}_{t\rightarrow s}, \cdots, \mathbf{w}^{(5)}_{t\rightarrow s}\}, I^{r}_t).
\end{aligned}
\end{equation}
%%%%%%%%%%%%%%%%%%%%%%

Similarly, $\SynthesisNet$ (see Eq.~\ref{eq:synthesisnet}) utilizes 
$[I_s, P_s]$ to encode source domain information, $[\hat{G}_t,P_t]$ to encode target domain information, $\{\mathbf{w}^{(0)}_{t\rightarrow s}, \mathbf{w}^{(1)}_{t\rightarrow s}, \cdots, \mathbf{w}^{(5)}_{t\rightarrow s}\}$ from stage-\RNum{1} for alignment, and $I^{r}_t$ to keep the background, hair and face of target image.  The output of $\SynthesisNet$ is the synthesized image $\hat{I}_t$:
%%%%%%%%%%%%%%%%%%%%%%
\begin{equation}
\label{eq:synthesisnet}
\begin{aligned}
\hat{I}_t=\SynthesisNet(&[I_s, P_s], [\hat{G}_t, P_t],\\
&\{\mathbf{w}^{(0)}_{t\rightarrow s}, \mathbf{w}^{(1)}_{t\rightarrow s}, \cdots, \mathbf{w}^{(5)}_{t\rightarrow s}\},I^{r}_t).
\end{aligned}
\end{equation}
%%%%%%%%%%%v%%%%%%%%%%%

Since the two networks share the similar inputs format and network structure, we elaborate the shared network structure below.
%in Section~\ref{subsection:method-stage2}.

%\subsubsection{Network Structure}
\vspace{0.1cm}
\noindent \textbf{Network Structure}\quad
As shown in Fig.~\ref{fig:GarmentNet} and \ref{fig:SynthesisNet}, our model relies on a source encoder $\Enc_{s}(\cdot)$ and a target encoder $\Enc_{t}(\cdot)$ to respectively generate multi-scale feature maps from source and target domains inputs $\mathit{IN}_s, \mathit{IN}_{t}$:

\begin{equation}
\begin{aligned}
\{\mathbf{f}^{(0)}_s, \cdots, \mathbf{f}^{(5)}_s\} = \Enc_{s}({\mathit{IN}_s}),\\
\{\mathbf{f}^{(0)}_t, \cdots, \mathbf{f}^{(5)}_t\} = \Enc_{t}({\mathit{IN}_t}).
\end{aligned}
\end{equation}
For GarmentNet, inputs are set to $\mathit{IN}_s=[G_s, P_s], \mathit{IN}_{t}=P_t$. For SynthesisNet, inputs are set to $\mathit{IN}_s=[I_s, P_s], \mathit{IN}_{t}=[\hat{G}_t, P_t]$.

We use six stacked strided convolutional layers to implement $\Enc_{t}(\cdot)$ and six stacked strided convolutional layers following seven residue blocks to implement $\Enc_{s}(\cdot)$. The additional residue blocks serve to increase feature representation capacity.

To perform spatial alignment, the source domain features $\mathbf{f}^{(l)}$ at all scales $l \in \{0,\dotsi ,5\}$ are inversely warped \cite{STN} to target domain using $\mathbf{f}^{(l)}_s$ and $\mathbf{w}^{(l)}_{t\rightarrow s}$ for layers $l\in\{1,\cdots,5\}$, formally:
\begin{equation}
\mathbf{f}^{(l)}_{s\rightarrow t} = \warp(\mathbf{f}^{(l)}_s;\mathbf{w}^{(l)}_{t\rightarrow s}).
\end{equation}

After spatial alignment, a U-Net fusion decoder is used for feature aggregation. However, instead of directly concatenating feature maps for aggregation, we propose a gated multiplicative attention module to filter the misaligned source domain features. 
%Specifically, let $\mathbf{f}_s$ and $\mathbf{f}_t$ denote the warped source feature and target feature at the same location, 
Specifically, the gated multiplicative attention filtering at scale $l$ is defined as:
%%%%%%%%%%%%%%%%%%%%%%
\begin{equation}
\label{eq:multiplicative_filter}
\mathbf{f}_{s\rightarrow t}^{(l)\prime} = \mathbf{f}^{(l)}_{s\rightarrow t} \odot \sigma(\mathbf{f}^{(l)\top}_{s\rightarrow t} \mathbf{W}^{(l)} \mathbf{f}^{(l)}_t),
\end{equation}
%%%%%%%%%%%%%%%%%%%%%%
where $\sigma(\cdot)$ represents the sigmoid function, $\odot$ represents element-wise multiplication and $\mathbf{W}^{(l)}$ is a learnable matrix that measures dot product similarities between $\mathbf{f}^{(l)}_s$ and $\mathbf{f}^{(l)}_t$ on to-be-learned linear space. The gated multiplicative attention filtering can be efficiently implemented on the 2-D feature maps using $1\times1$ convolution, element-wise multiplication and summation. Please refer to Appendix~\ref{appendix:filtering} for details. Building on top of the gated multiplicative attention filtering operation, our decoder uses the following equations to generate the aggregated feature maps $\mathbf{f}^{(l)}_{dec}$:
\begin{equation}
\label{eq:decoder}
\begin{aligned}
\mathbf{f}^{(0)}_{dec} &= \Deconv([\mathbf{f}_{s\rightarrow t}^{(0)\prime}; \mathbf{f}^{(0)}_{t}]),\\
\mathbf{f}^{(l)}_{dec} &= \Deconv([\mathbf{f}^{(l-1)}_{dec}; \mathbf{f}_{s\rightarrow t}^{(l)\prime}; \mathbf{f}^{(l)}_{t}]),l\in\{1,\cdots,5\}.\\
\end{aligned}
\end{equation}

Afterwards, our network simultaneously generates foreground content $\mathit{fg}$ along with a mask $M$ that ranges from $0$ to $1$  to avoid changing the residue content of the target $r_t$. Specifically,$\mathbf{f}^{(5)}_{dec}$ is passed to two independent convolutional layers to respectively generate foreground content $\mathit{fg}$ and a corresponding foreground mask $\mathit{M}$:
\begin{equation}
\label{eq:mask}
\begin{aligned}
\mathit{fg} &= \Conv(\mathbf{f}^{(5)}_{dec}), \\
\mathit{M} &= \Conv(\mathbf{f}^{(5)}_{dec}).
\end{aligned}
\end{equation}

Finally, the output content $\mathit{out}$ is generated by alpha-blending the foreground content $\mathit{fg}$ with the residue content $r$:
\begin{equation}
\begin{aligned}
\mathit{out} = \mathit{M} \odot \mathit{fg} +  (1-\mathit{M}) \odot r.
\end{aligned}
\end{equation}

For GarmentNet, softmax function is applied after $\mathit{out}$ to generate the garment parsing, i.e. $\hat{G}_s=\softmax(\mathit{out})$. For SynthesisNet, tanh function is applied after $\mathit{out}$ to generate the normalized image, i.e. $\hat{I}_s=\tanh(\mathit{out})$.

%%%%%%%%%%%%%%%%%%%%%%
\vspace{0.1cm}
\noindent \textbf{Training Objective}\quad
For GarmentNet training, we use the cross entropy loss between the target garment $G_{t}$ and prediction $\hat{G}_t$:
%%%%%%%%%%%%%%%%%%%%%%
\begin{equation}
\label{eq: cross_entropy_loss}
\mathcal{L}_{\text{GarmentNet}} = - \sum_{i,j} \sum_{n} \left(G_{t}\right)_{i,j,n} \log( (\hat{G}_{t})_{i,j,n}),
\end{equation}
where $i,j$ enumerate pixel positions and $n$ enumerates channals of garment parsing.

For SynthesisNet training, we use a combination of $\ell_1$ pixel domain loss, VGG feature loss, texture loss, and GAN loss. The training objective is represented as:
%%%%%%%%%%%%%%%%%%%%%%
\begin{align}
\label{eq: total_loss}
\mathcal{L}_{\text{SynthesisNet}} = & \lambda_{1} \mathcal{L}_1  + \lambda_{2} \mathcal{L}_{\text{VGG}} \nonumber \\
& + \lambda_{3} \mathcal{L}_{\text{texture}} + \lambda_{4} \mathcal{L}_{\text{GAN}},
\end{align}
%%%%%%%%%%%%%%%%%%%%%%
where $\mathcal{L}_1=\norm{\hat{I}_t - {I}_{t}}_1$ computes the $\ell_1$ differences between the synthesized image and the ground-truth, $\mathcal{L}_{\text{VGG}} = \norm{\mathbf{f}_{\text{VGG}}(\hat{I}_t) - \mathbf{f}_{\text{VGG}}({I}_{t})}_1$ computes feature map differences on the \texttt{relu4\_2} layer of the VGG network of the two image. Similar to Eq.~\ref{eq:struct}, $\mathcal{L}_{\text{texture}}=\norm{ \mathbf{G}\left(\mathbf{f}_{\text{VGG}}(\hat{I}_t)\right)-\mathbf{G}\left(\mathbf{f}_{\text{VGG}}(I_t)\right) }_1$ (Eq.~\ref{eq:struct}) computes the texture-level differences of the two images, and $\mathcal{L}_{\text{GAN}}={(D(I_t)-1)}^2 + {D(\hat{I}_t)}^2$ measures how well the synthetic image can fool a trained discriminator $D(\cdot)$. Similar to CycleGAN~\cite{cyclegan}, we use least-square distance~\cite{lsgan} rather than negative log likelihood to compute the  $\mathcal{L}_{\text{GAN}}$, whereas the discriminator is implemented using the PatchGAN architecture~\cite{pix2pix} with spectrum normalization~\cite{sn}. The hyper-parameters $\lambda_1,\lambda_2,\lambda_3,\lambda_4$ are set to $\lambda_1= 1.0, \lambda_2= 0.1,\lambda_3=0.002,\lambda_4=0.5$ respectively in our experiments. 

Additionally, we use a similar augmentation-based self-supervision strategy as described in Sec.~\ref{subsection:method-stage1} to regularize SynthesisNet. 
% Specifically, we use the following update rules to transform the original data before one iteration of SynthesisNet training: 
% \begin{equation}
% \begin{aligned}
% \epsilon &\sim U(0,1),\\
% P_t &\gets \Aug(P_s,\theta),\text{if $\epsilon<0.25$}\\
% G_t &\gets \Aug(G_s,\theta),\text{if $\epsilon<0.25$}\\
% I_t &\gets \Aug(I_s,\theta),\text{if $\epsilon<0.25$}.
% \end{aligned}
% \end{equation}
During training, $25\%$ percent of the source domain samples come from the augmented target domain samples to help SynthesisNet to learn from simple tasks first.

\begin{table*}[]
	\centering
	\caption{Quantitative comparison of different methods in terms of both the masked SSIM/msSSIM/Inception Score (IS) and the Learned Perceptual Image Patch Similarity (LPIPS) at $256 \times 256$ and $128 \times 128$ resolution. Higher scores are better for metrics with uparrow ($\uparrow$), and vice versa.}
	\begin{tabular}{ |l|c|c|c|c|c|c|c|c}
		\toprule
		\hline
		Methods& 								 SSIM-128$\uparrow$ &     msSSIM-128$\uparrow$  & 	SSIM$\uparrow$ &      msSSIM$\uparrow$  & 	IS-128$\uparrow$  & IS$\uparrow$ 			& LPIPS$\downarrow$		& LPIPS-128$\downarrow$\\
		\hline
		PG2~\cite{pg2}  &                       0.864& 	0.911& 0.857& 0.891&      3.455 $\pm$ 0.226 & 4.266 $\pm$ 0.371			&0.192&  0.190	\\
		BodyROI7~\cite{disentangle_gan} &       0.842& 		0.882& 0.837& 		0.865&      3.282 $\pm$ 0.173 & 3.855 $\pm$ 0.158			&0.193&  0.201\\
		DSCF~\cite{deform_gan}&                 0.856& 		0.902& 0.851& 		0.884&      3.458 $\pm$ 0.198 & 4.226 $\pm$ 0.326			&0.159&  0.157\\
		Vunet~\cite{vunet}&                     0.822&0.830& 0.827& 0.827&   	3.424 $\pm$ 0.143 & 4.176 $\pm$ 0.320			&0.226&  0.258\\
		Soft-gate~\cite{softgated} &            0.860& 		0.908& 0.853& 		0.888&      3.270 $\pm$ 0.219 & 3.868 $\pm$ 0.387			&0.140&  0.135\\
		IF~\cite{intrinsic_flow} &              \bf0.877& \bf0.926& \bf0.865& \bf0.906&     3.262 $\pm$ 0.293 & 3.809 $\pm$ 0.360			&0.128&  0.128\\
		\hline
		Ours			&0.854& 		0.905& 0.848& 		0.884& 		{3.540 $\pm$ 0.294} & {4.197 $\pm$ 0.291} &\bf0.124&  \bf0.124\\
		Ours-kp   	&0.831& 		0.870& 0.831& 		0.852& {\bf3.646 $\pm$ 0.285} & {\bf4.295 $\pm$ 0.296}			&0.163&  0.169\\
		\hline
        \bottomrule
	\end{tabular}
	\label{Table:main_experiment_256}
\end{table*}

\begin{figure*}[t]
	\centering
	\includegraphics[width=1.0\linewidth]{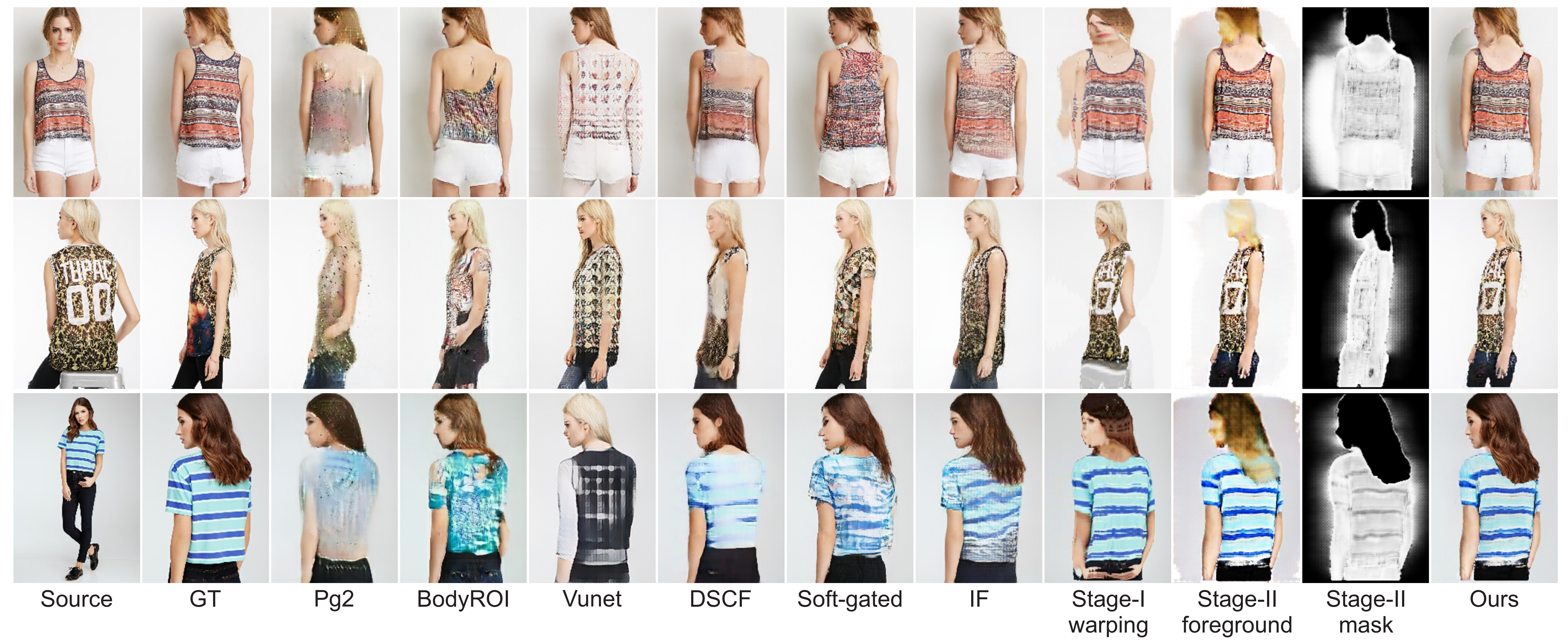}
	\caption{Comparison with the state-of-the-art approaches. 
		The last four columns depict the warped source image, fore ground prediction in stage-\RNum{2}, mask prediction in stage-\RNum{2}, and our final output. In comparison, our method clearly produces the most visually plausible and pleasing effects.}
	\label{fig:compare_main}
\end{figure*}

%\begin{table*}[t]
%	\centering
%	\begin{tabular}{ l|c|c|c|c|c|c}
%        \toprule
%        \hline
%         {SyntnesisNet}& 			\multirow{2}{*}{SSIM} &    
%         \multirow{2}{*}{mask-SSIM} &
%         \multirow{2}{*}{msSSIM} &
%         \multirow{2}{*}{mask-msSSIM} &
%         \multirow{2}{*}{IS} &
%         \multirow{2}{*}{mask-IS} \\
%       {training schemes} & & & &  &\\
%
%        \hline
%		
%       w/o flow&                            0.816&      0.868  &    0.832&          0.873&          2.848 $\pm$ 0.196&      {\bf3.866 $\pm$ 0.358} \\
%        w/o att&                            0.830&      0.880  &    0.842&          0.882&          2.806 $\pm$ 0.179&      3.621 $\pm$ 0.262\\
%       w/o semi&                            0.832&      0.882   &   0.844&          0.883&          2.843 $\pm$ 0.197&      3.702 $\pm$ 0.276\\
%          Full model&                             {\bf0.834}& {\bf0.885}  & {\bf0.845}&   {\bf0.885}&     {\bf2.850 $\pm$ 0.224}& 3.718 $\pm$ 0.381\\
%		\hline                                                         
%	\bottomrule
%	\end{tabular}
%	\caption{Performance comparison of different SynthesisNet training schemes in terms of SSIM, msSSIM and their masked versions.}
%	\label{tab:compare_model_ablation}
%\end{table*}

\section{Experiments}
\label{sec:experiment}
\subsection{Dataset}
We train and evaluate our method on the DeepFashion~\cite{deepfashion} dataset, which contains 52,712 person images of sizes $256 \times 256$.
Images that only contain trousers are removed using DensePose~\cite{densepose}, resulting in 40,906  valid images.  We randomly divide the dataset into 68,944 training pairs and 1,000 testing pairs. Additionally, we evaluate our DeepFashion trained model on other datasets to understand how well our model can generalize to unseen poses, clothing styles or background.

As detailed in Section~\ref{subsection:notation}, pose representation are generated using DensePose, while garment representation are generated using the method of~\cite{LIP}. Finally, we additionally uses keypoint heatmap \cite{pg2} as pose representation to test our algorithm.

%%%%%%%%%%%%%
\begin{figure}[]
\centering
\includegraphics[width=1.0\linewidth]{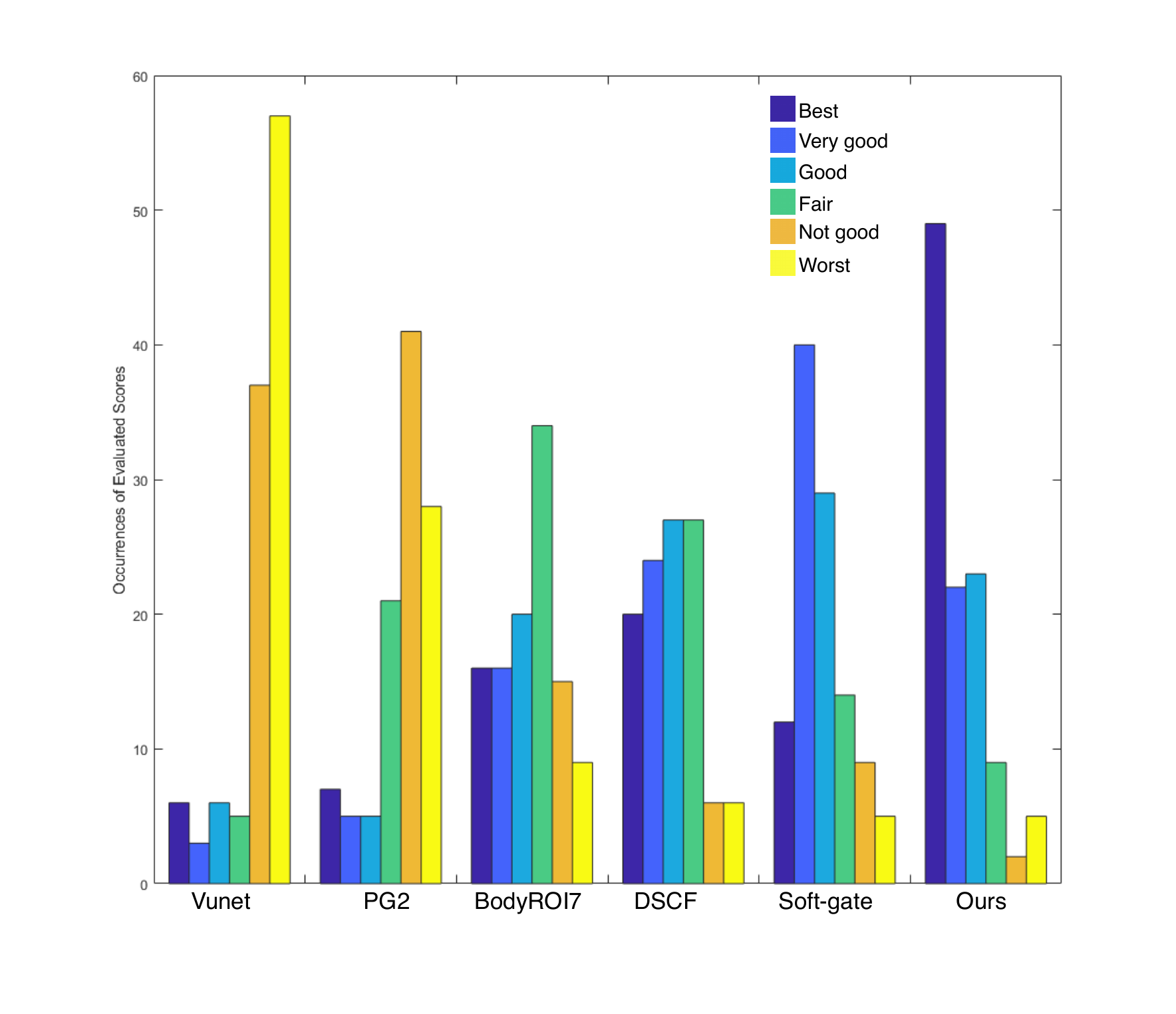}
\caption{Subjective quality assessment of different algorithms. For each algorithm, the bar depicts the number of occurrences of scores, while blue to yellow colors represent the scores from the best to the worst.}
\label{fig:user_study}
% \vspace{-3mm}
\end{figure}
\subsection{Implementation Details}
In Stage-\RNum{1} and Stage-\RNum{2}, we set the learning rate to $0.0001$ for the flow estimator and the generator. Following~\cite{sn}, the learning rate for the  discriminator is $0.0004$. We adopt Adam~\cite{adam} optimizer ($\beta_1=0.9$ and $\beta_2=0.999$) in all experiments. Random cropping, affine transformation and flipping are used to augment data. The flow estimator, GarmentNet and SynthesisNet are trained for $20$, $20$ and $40$ epochs, respectively. 

Since our approach can adopt keypoint heatmap \cite{pg2} as pose representation by simply altering $P_s, P_t$, we additionally train our model using the key point representation while maintaining other inputs unchanged.

\vspace{-2mm}
\subsection{Quantitative Evaluation}
To quantitatively evaluate the synthesis results, low-level metrics like Structural Similarity (SSIM)~\cite{ssim}, Multi-scale Structural Similarity (MS-SSIM)~\cite{ms-ssim} and perceptual-level metrices like Inception Score (IS)~\cite{is} and the Perceptual Image Patch Similarity Distance (LPIPS)~\cite{lpips} are measured on different approaches, including PG2~\cite{pg2}, BodyROI~\cite{disentangle_gan}, Vunet~\cite{vunet}, DSCF~\cite{deform_gan}, Soft-gated GAN (Soft-gate)~\cite{softgated} and Intrinsic Flow (IF)~\cite{intrinsic_flow}. For LPIPS, we use the linearly calibrated Alex model, please refer to~\cite{lpips} for details.
Since our approach relies on the background information, we report the masked version of all the metrices for fair comparisons. The masks are generated by running \cite{LIP} to exclude background, hair, and face region. We additionally test all the metrics at resolution $128 \times 128$ to measure similarities at a global scale. 

From Table \ref{Table:main_experiment_256}, our method (\textit{ours}) substantially outperforms the remaining methods in IS-based measurements and LPIPS distances, as our texture-preserving flow is able to preserve texture patterns form source images. In terms of the low-level SSIM-based measurements, our method achieves competitive performance in comparison with the other approaches. When trained using keypoint heatmap (\textit{ours-kp}), we observe similar high IS scores for both models and better LPIPS scores for our model. It suggests both models (\textit{ours} and \textit{ours-kp}) preserve realistic texture. However, with the help of the DensePose pose representation, our model (\textit{ours}) generates better global shape.

\subsection{Qualitative Evaluation}
We conduct a subjective assessment to evaluate our method qualitatively. Specifically, we ask 15 subjects to rank image qualities among the 6 algorithms (\cite{vunet,pg2,disentangle_gan,deform_gan,softgated} and ours). The subjects are instructed to rank the six images, based on the realism of the generated garments as well as global garment structures. The subjects are then asked to provide a score from $1$ to $6$ for each image, representing best quality to worst quality,  respectively. We plot the ranking histogram of different algorithms in Fig. \ref{fig:user_study}. From the figure, our method is most frequently chosen as the best due to structurally consistent texture. DSCF \cite{deform_gan} achieves the second place due to its ability to maintain texture structure from the source image using rigid transformations. The qualitative results of different approaches as well as the warped source image and foreground/mask prediction from stage-\RNum{2} are shown in Fig.~\ref{fig:compare_main}. It can be noticed that the existed approaches generate blurry results or incorrect textures. By contrast, our method can preserve texture details from source images. Notably, our approach generates better warping results in comparison with IF, especially under large pose changes.

\begin{table*}[]
	\centering
	\caption{Quantitative comparison of different flow training schemes and  SynthesisNet training schemes in terms of both the masked SSIM/msSSIM/Inception Score (IS) and the Learned Perceptual Image Patch Similarity (LPIPS) at $256 \times 256$ and $128 \times 128$ resolution. Higher scores are better for metrics with up arrows ($\uparrow$), and vice versa.}
	\begin{tabular}{ l|c|c|c|c|c|c|c|c}
		\toprule
		\hline
		Flow training schemes& 		SSIM-128$\uparrow$ &     msSSIM-128$\uparrow$  & 	SSIM$\uparrow$ &      msSSIM$\uparrow$  & 	IS-128$\uparrow$  & IS$\uparrow$ 	& LPIPS$\downarrow$		& LPIPS-128$\downarrow$\\
		\hline
		w/o multi-scale	&0.822& 0.853& 0.825& 0.839&		\bf{4.115 $\pm$ 0.211} & \bf{4.689 $\pm$ 0.327}	&0.240&  0.240\\
		w/o texture		&\bf0.837& 0.880& \bf0.837& 0.861&		3.843 $\pm$ 0.246 & 4.204 $\pm$ 0.245			&0.217&  0.217\\
		w/o semi		&0.835& 0.880& 0.834& 0.861&		3.978 $\pm$ 0.348 & 4.412 $\pm$ 0.223			&0.196&  0.196\\
		full training scheme&0.836& \bf0.882& 0.835& \bf0.863& 3.934 $\pm$ 0.274 & 4.404 $\pm$ 0.331&\bf0.193&  \bf0.193\\
		\hline                     
		\hline                     
		SynthesisNet training schemes& 		SSIM-128$\uparrow$ &     msSSIM-128$\uparrow$  & 	SSIM$\uparrow$ &      msSSIM$\uparrow$  & 	IS-128$\uparrow$  & IS$\uparrow$ 			& LPIPS$\downarrow$		& LPIPS-128$\downarrow$\\
		\hline
		w/o flow	&0.849& 0.898& 0.844& 0.877&		3.421 $\pm$ 0.177 & 3.952 $\pm$ 0.291			&0.141&  0.141\\
		w/o att		&0.853& 0.904& 0.848& 0.883&		3.391 $\pm$ 0.161 & 3.946 $\pm$ 0.374			&0.128&  0.128\\
		w/o semi	&0.851& 0.903& 0.846& 0.882&		3.480 $\pm$ 0.273 & 3.995 $\pm$ 0.333			&0.128&  0.128\\
		full model	&\bf0.854& \bf0.905& \bf0.848& \bf0.884& 	\bf{3.540 $\pm$ 0.294} & \bf{4.197 $\pm$ 0.291}			&\bf0.124&  \bf0.124\\
		\hline               
		\bottomrule
	\end{tabular}
	\label{tab:compare_ablation}
\end{table*}

% \begin{table*}[]
% 	\centering
%     \caption{Quantitative comparison of different SynthesisNet training schemes in terms of both the masked SSIM/msSSIM/Inception Score (IS) and the Learned Perceptual Image Patch Similarity (LPIPS) at $256 \times 256$ and $128 \times 128$ resolution. Higher scores are better for metrics with uparrow ($\uparrow$), and vice versa.}
% 	\begin{tabular}{ l|c|c|c|c|c|c|c|c}
% 		\toprule
% 		\hline
% 		Methods& 		SSIM-128$\uparrow$ &     msSSIM-128$\uparrow$  & 	SSIM$\uparrow$ &      msSSIM$\uparrow$  & 	IS-128$\uparrow$  & IS$\uparrow$ 			& LPIPS$\downarrow$		& LPIPS-128$\downarrow$\\
% 		\hline
% 		w/o flow	&0.849& 0.898& 0.844& 0.877&		3.421 $\pm$ 0.177 & 3.952 $\pm$ 0.291			&0.141&  0.141\\
% 		w/o att		&0.853& 0.904& 0.848& 0.883&		3.391 $\pm$ 0.161 & 3.946 $\pm$ 0.374			&0.128&  0.128\\
% 		w/o semi	&0.851& 0.903& 0.846& 0.882&		3.480 $\pm$ 0.273 & 3.995 $\pm$ 0.333			&0.128&  0.128\\
% 		full model	&\bf0.854& \bf0.905& \bf0.848& \bf0.884& 	\bf{3.540 $\pm$ 0.294} & \bf{4.197 $\pm$ 0.291}			&\bf0.124&  \bf0.124\\
% 		\hline                                                              
% 		\bottomrule
% 	\end{tabular}
% 	\label{tab:compare_model_ablation}
% \end{table*}

\subsection{Ablation Study}
\begin{figure}[]
	\centering
	\includegraphics[width=1.0\linewidth]{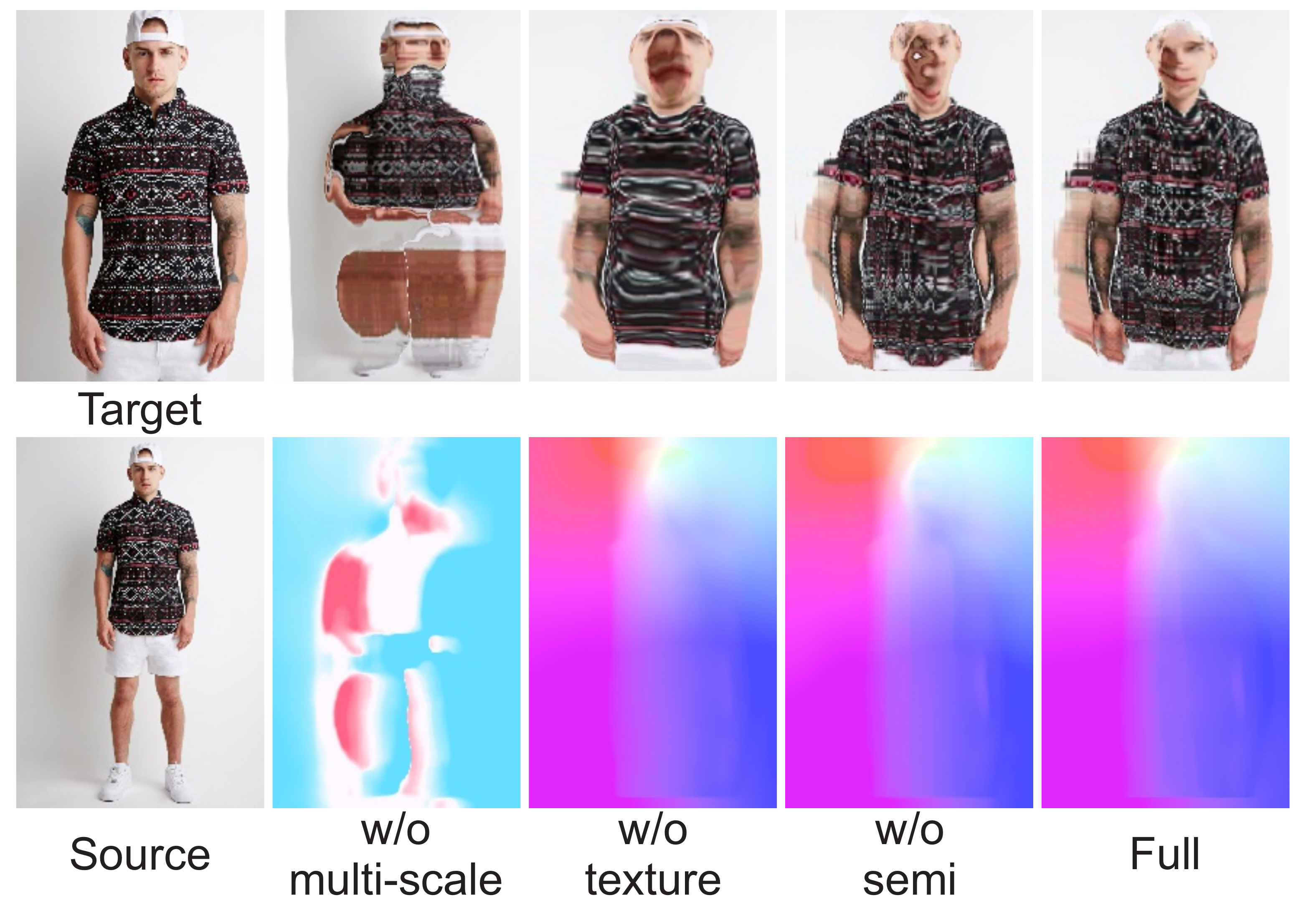}
	\includegraphics[width=1.0\linewidth]{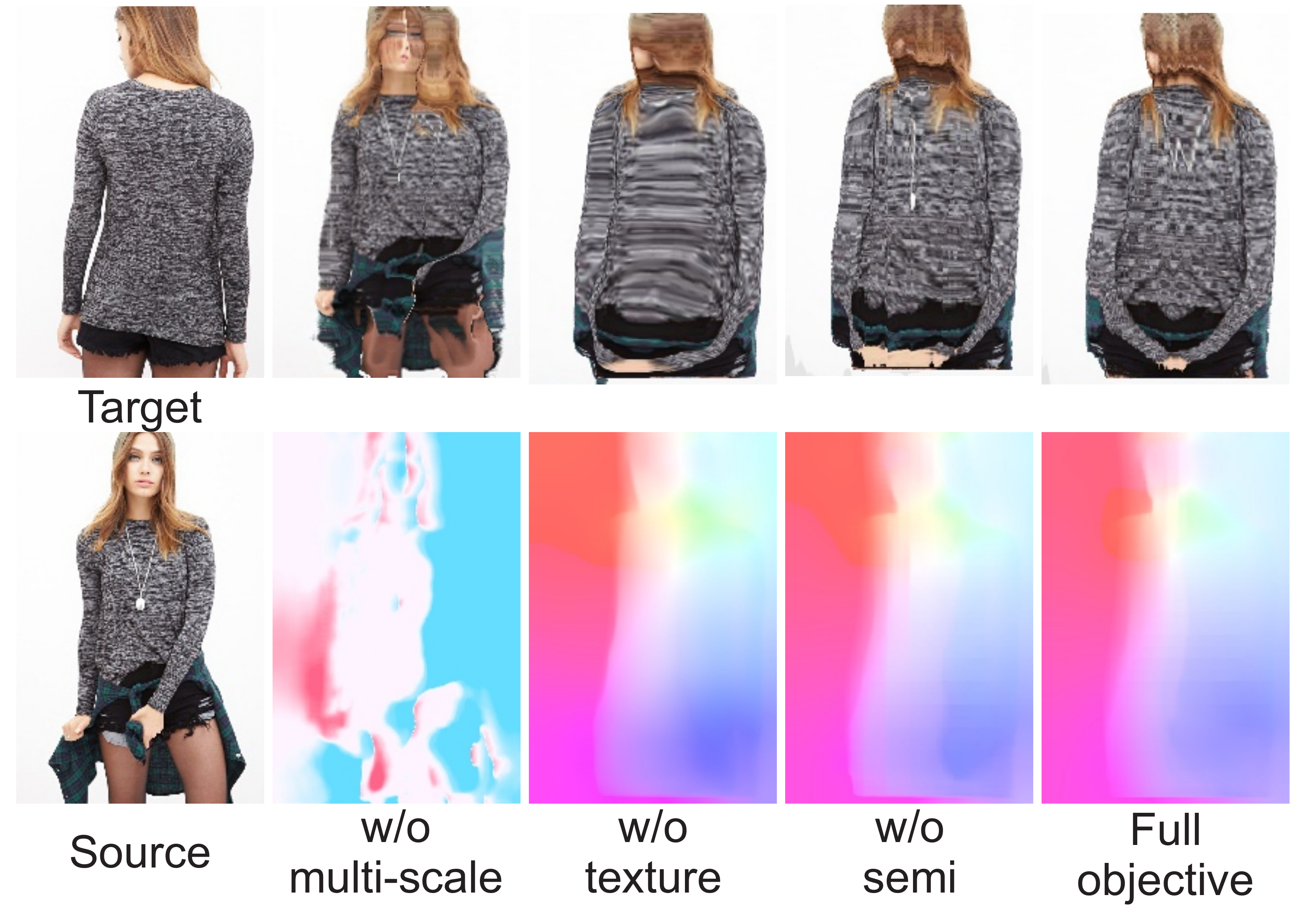}
	\caption{Comparisons of different unsupervised flow training schemes. Our full flow training objective (Eq.~\ref{eq:stage1_loss}) generates more visually plausible and pleasing textures and more consistent flow.}
	\label{fig:compare_flow}
\end{figure}

\noindent \textbf{Unsupervised Flow Training} \quad
To evaluate the effectiveness of each component in the unsupervised flow training scheme, we separately train three variants of the proposed flow estimators: i) \textit{w/o multi-scale}, only computing loss at the finest scale, ii) \textit{w/o texture}, removing texture loss $\mathcal{L}_{texture}$, and iii) \textit{w/o semi}, removing the augmentation-based self supervision. Table \ref{tab:compare_ablation} compares the three models with our full model by computing the SSIM, IS, and LPIPS-based scores of the inversely warped images using the trained flow at the finest scale. The inversely warped image is also visualized in Fig.\ref{fig:compare_flow}. It is observed that our full model outperforms \textit{w/o semi} and \textit{w/o multi-scale} in terms of LPIPS scores. It is consistent with the visualization from Fig. \ref{fig:compare_flow}, showing that our full model can generate flow with more visually plausible and pleasing details. The \textit{w/o multi-scale} performs well in IS scores, and it is possibly because \textit{w/o multi-scale} tends to retain the realistic original source image. However, \textit{w/o multi-scale} does not preserve the semantics of the target pose. In terms of SSIM-based measurement, the full flow training scheme achieves the best ms-SSIM scores, suggesting that the full model is better at preserving global structures.

\begin{figure}[]
	\centering
	\includegraphics[width=1.0\linewidth]{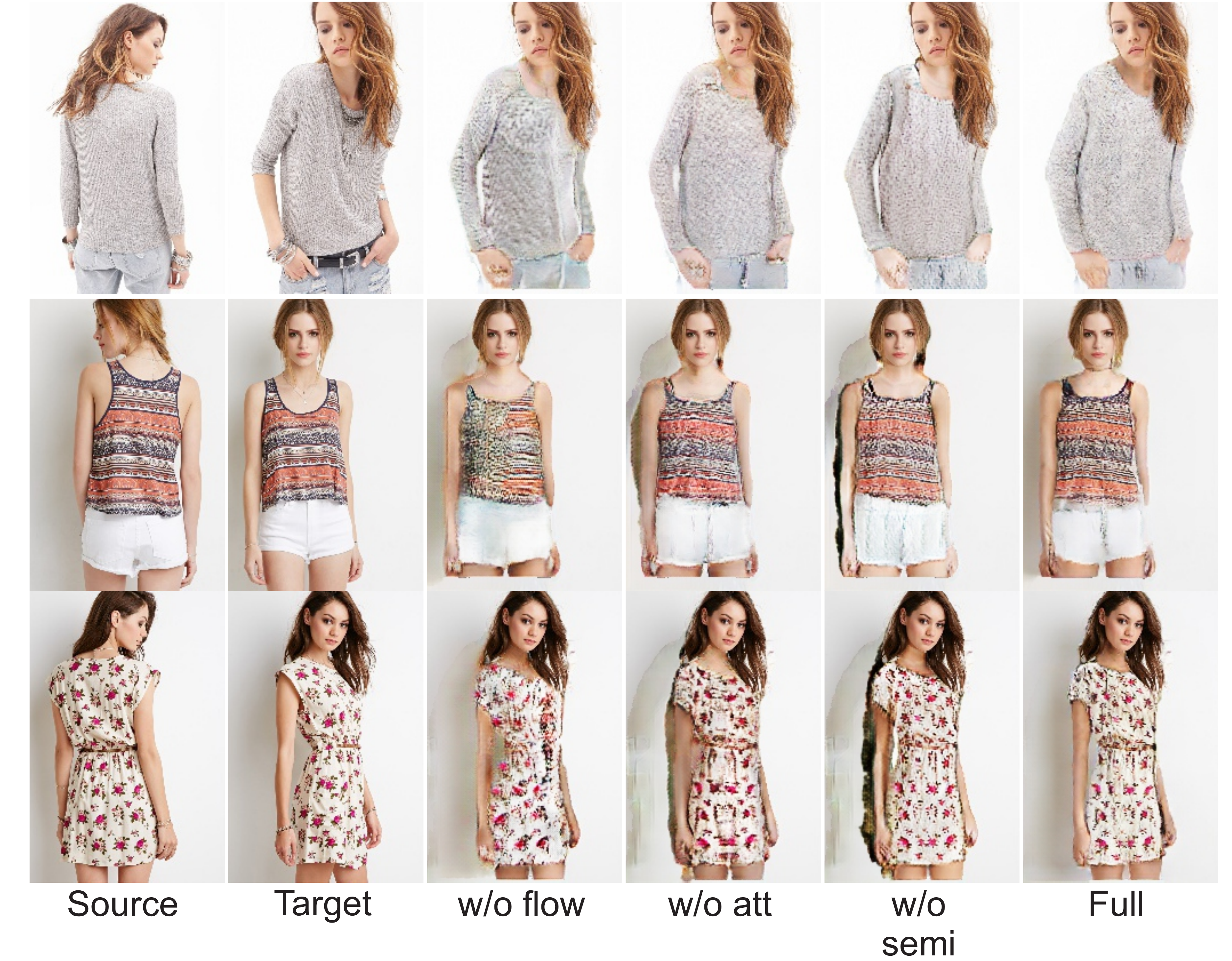}
	\caption{Visual comparisons of different SynthesisNet training schemes. Our full model generates more visually plausible and pleasing texture details with more coherent global structures.}
	\label{fig:compare_model_ablation}
\end{figure}

\vspace{0.2cm}
\noindent \textbf{SynthesisNet Design} \quad
To evaluate the effectiveness of different components in training SynthesisNet, an ablation study is performed in the following ways: i) we remove the flow estimator for alignment, resulting in \textit{w/o flow}, a UNet-like structure that does not perform feature alignment, ii) we replace the gated multiplicative attentive fusion modules with concatenation operations, which is called \textit{w/o att}, iii) we replace the semi-supervised data generation scheme with only the supervised data, which is called \textit{w/o semi}. Table \ref{tab:compare_ablation} compares the qualitative scores in terms of SSIM, ms-SSIM, IS and their masked versions. From the table, we observe that the SSIM-based performances substantially deteriorate without the flow-based alignment module. Meanwhile, the gated multiplicative attentive fusion helps to improve the inception scores of the generated images. Also,  semi-supervised training improves performance marginally. Visualization is also shown in Fig.~\ref{fig:compare_model_ablation}. From the figure, we observe that our full model is able to retain the global structure due to flow-based alignment. Comparing \textit{w/o att} and \textit{full}, we see that with the gated multiplicative attention module, our model generates globally consistent texture details. 

\vspace{-2mm}
\subsection{Generalization}

\begin{figure*}[]
	\centering
	\includegraphics[width=.8\linewidth]{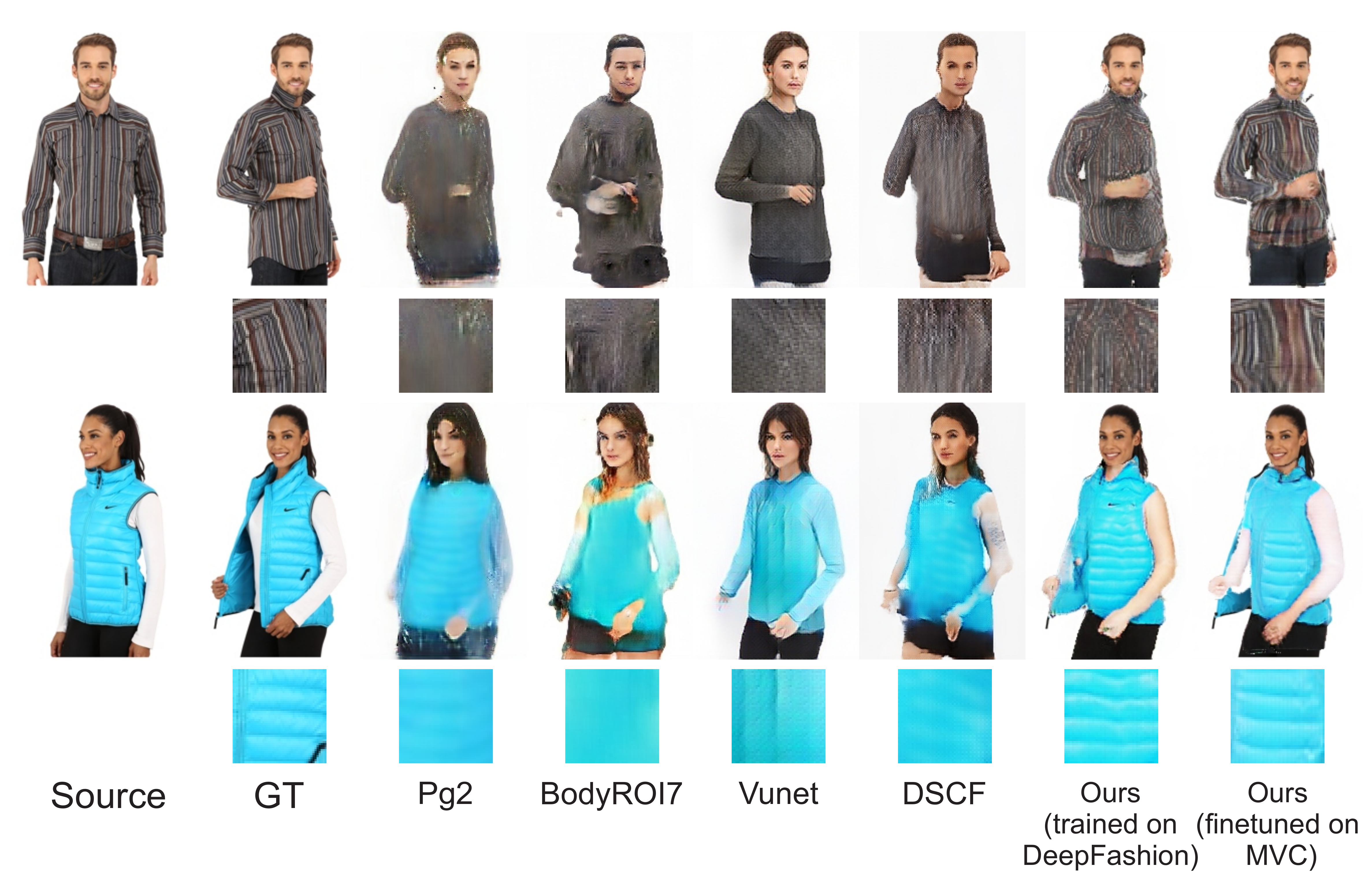}
	\caption{Comparison with the state-of-the-art approaches on the MVC dataset. Patches are zoomed in to visualize detailed textures. The last two columns depict our DeepFashion trained model and our MVC finetuned model.}
	\label{fig:compare_main_MVC}
\end{figure*}

To understand the generalization ability of our trained model and how well our model can perform on real-world datasets, we evaluate  our trained model on three additional datasets:

\noindent \textbf{Multi-view Clothing dataset} \quad
The Multi-view  Clothing dataset (MVC)  \cite{MVC} contains 161,260 person images and 645,040 pairs in total. We report the results on the MVC dataset using various models that are trained on the DeepFashion dataset. We also report the performance of our finetuned model using 120,000 pairs selected from the MVC training set. Table~\ref{Table:main_experiment_MVC_256} shows the evaluation of our approach in comparison to other approaches. The generated new-person images are visualized in Fig.~\ref{fig:compare_main_MVC}.

\begin{table*}[]
	\centering
	\caption{Quantitative comparison of various approaches on the MVC dataset using the models trained on the DeepFashion dataset.  Performances are measured in terms of the masked SSIM/msSSIM/IS scores at $256 \times 256$ resolution and $128 \times 128$ resolution. Higher scores are better for metrics with up arrows ($\uparrow$), and vice versa. Top two scores are in bold.}
	\begin{tabular}{ l|c|c|c|c|c|c}
		\toprule
		\hline
		Methods& 								 	SSIM$\uparrow$ & SSIM-128$\uparrow$  & msSSIM$\uparrow$  & msSSIM-128$\uparrow$  & 	IS$\uparrow$  & IS-128$\uparrow$\\
		\hline
		PG2~\cite{pg2}  &                           0.817&          0.806&          0.851&      0.840&          3.401 $\pm$ 0.269 & 3.662 $\pm$ 0.361 \\
		BodyROI7~\cite{disentangle_gan} &           0.798&          0.792&          0.828&      0.823&          3.043 $\pm$ 0.250 & 3.039 $\pm$ 0.152 \\
		DSCF~\cite{deform_gan}&                     0.816&          0.810&          0.846&      0.841&          3.358 $\pm$ 0.229 & 3.151 $\pm$ 0.229 \\
		Vunet~\cite{vunet}&                         0.806&          0.794&          0.840&      0.833&          3.294 $\pm$ 0.190 & 2.871 $\pm$ 0.222 \\
		\hline
		Ours   &                        {\bf0.836}& {\bf0.839}&     {\bf0.857}& {\bf0.853}&          {\bf3.603 $\pm$ 0.300} & {\bf3.451 $\pm$ 0.426} \\
		Ours-Finetuned&                             {\bf0.839}&  {\bf0.840}&    {\bf0.863}& {\bf0.859}&          {\bf3.737 $\pm$ 0.415} & {\bf3.365 $\pm$ 0.273} \\
		\hline                                                              
		\bottomrule
	\end{tabular}
	\label{Table:main_experiment_MVC_256}
\end{table*}

\noindent \textbf{Amazon Fashion Video Data} \quad
We evaluate our approach on a set of online video data. Specifically, we crawl clothing item demo videos from the Amazon Fashion website. The initial frame from various source video is used as the source images to synthesize each frame from the target video. The synthesized videos are shown in the supplementary materials. In Fig.~\ref{fig:amazon_demo}, the top row shows the target video, while the resting rows show the synthesized video with different clothing styles from source images. As demonstrated in Fig.~\ref{fig:amazon_demo}, our approach generates temporal-consistent frames with distinctive texture details, suggesting that our method can effectively generalize to unseen poses and clothing styles.

\begin{figure*}[t]
	\centering
	\includegraphics[width=0.8\linewidth]{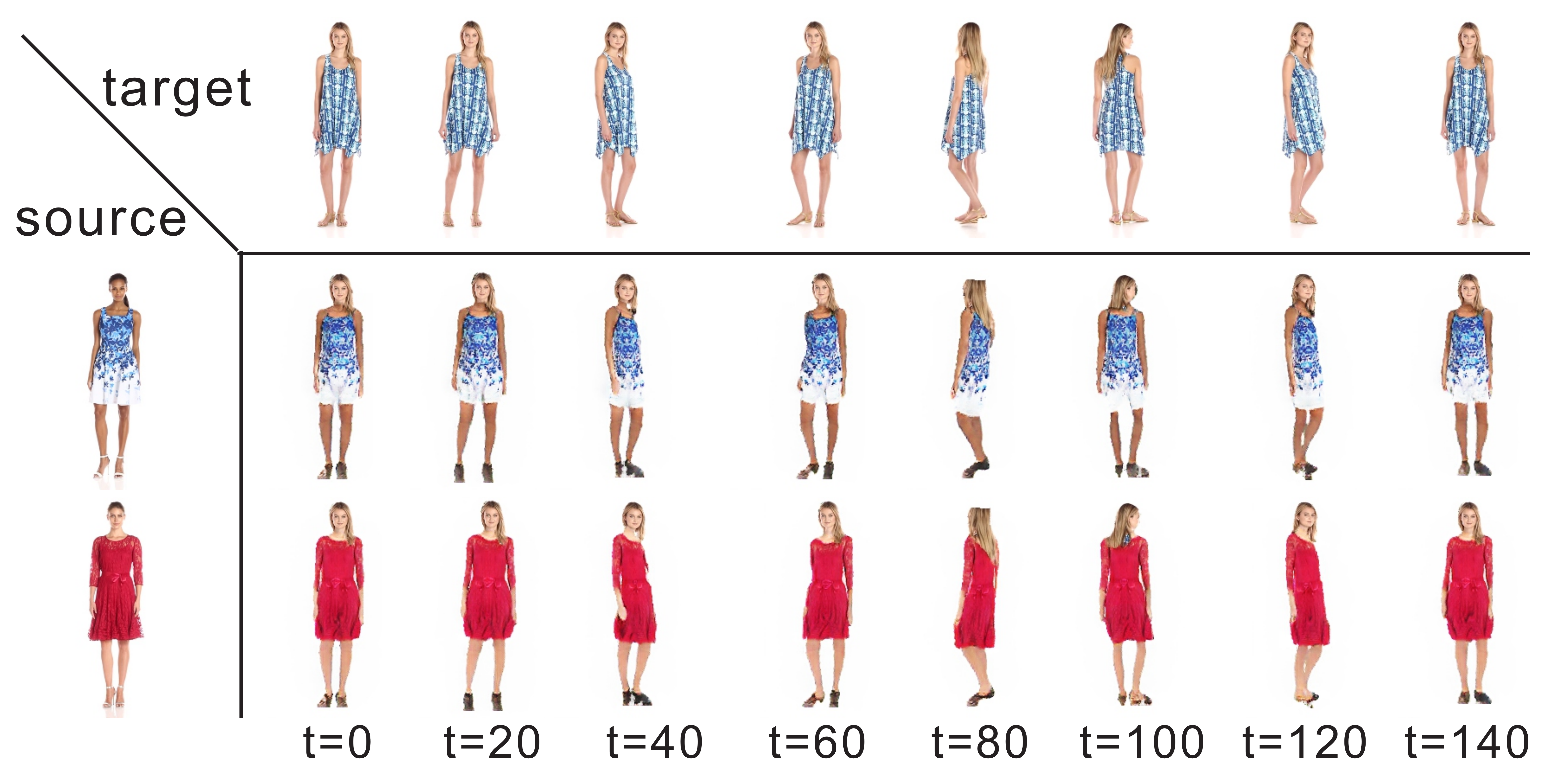}
	\caption{Garment transfer on the Amazon Fashion videos. The top row shows the target frames, while the resting rows show the synthesized frames. The horizontal axis represents the time step. Our approach can generate temporally consistent frames with distinctive texture details.}
	\label{fig:amazon_demo}
\end{figure*}

\begin{figure*}[]
	\centering
	\includegraphics[width=0.8\linewidth]{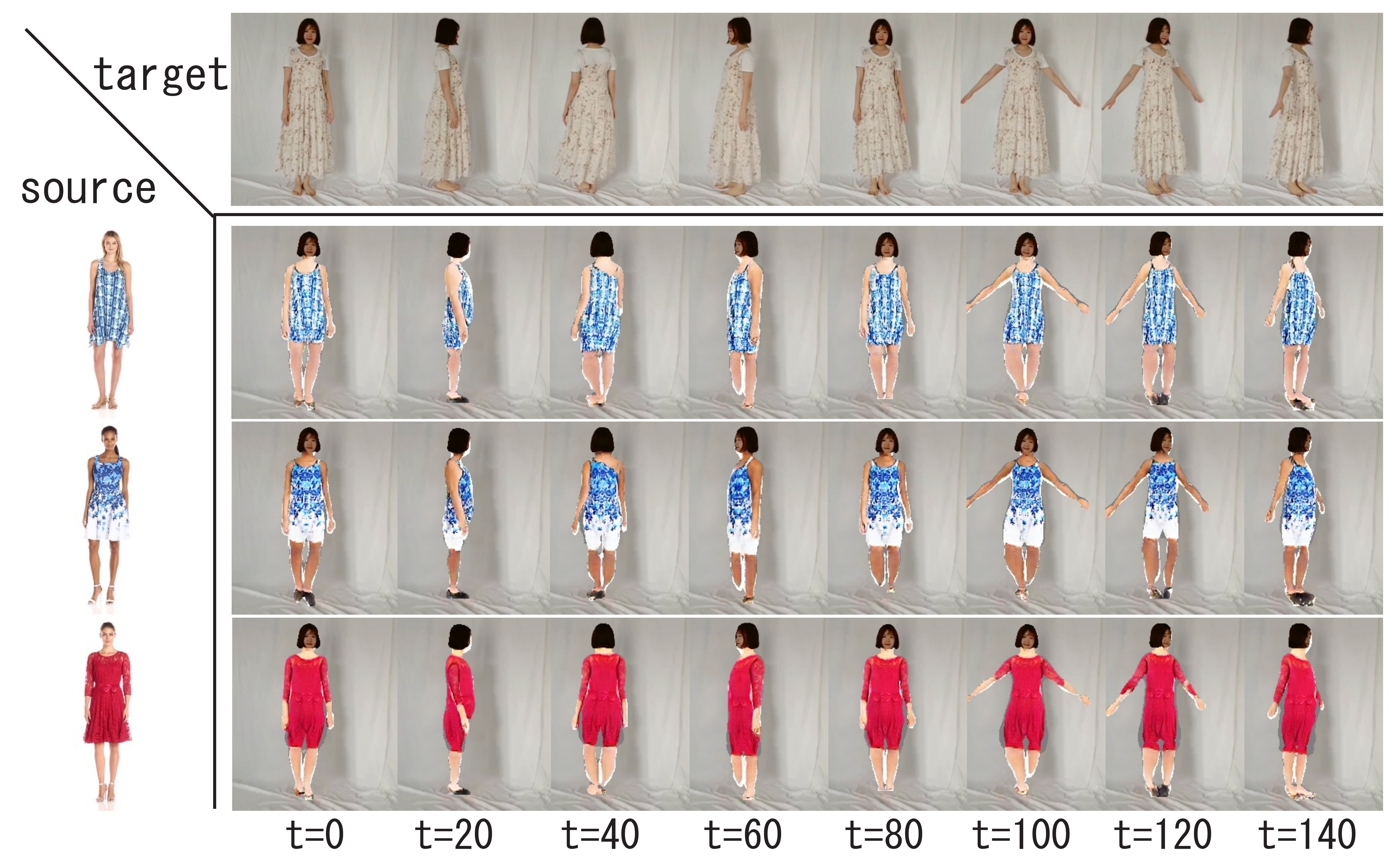}
	\caption{Garment transfer on our self-collected real-world videos. The top row shows the target frames, while the remaining rows show the synthesized frames. The horizontal axis represents the time step. Our approach can generate temporally consistent frames with distinctive texture details.}
	\label{fig:real_demo}
\end{figure*}

\noindent \textbf{Garment transfer to real person} \quad
To examine the applicability of our approach in real-world scenes, we collect videos of people in real scenes with various poses using a typical smartphone. Fig.~\ref{fig:real_demo} visualizes consecutive frames of our captured video and our transferred video, showing that our approach can generate visually plausible and pleasing new clothing styles under challenging real-world environments.

\vspace{-1mm}
\section{Conclusion}
\label{sec:conlusion}
To better model person appearance transformation for pose-guided synthesis, we propose an unsupervised pose flow learning scheme that learns to transfer appearance from target images. Furthermore, we propose a texture preserving objective and an augmentation-based self-supervision scheme which are shown to be effective for learning appearance-preserving pose flow. Based on the learned pose flow, we propose a coarse-to-fine synthesis pipeline using a carefully designed network structure for multi-scale feature domain alignment. To address the misalignment issue, we propose a gated multiplicative attention module. 
In addition, masking layers are used to preserve target identities and background information. Experiments on the DeepFasion, MVC, and other real-world datasets have validated the effectiveness and robustness of our approach.

% In this paper, we present an unsupervised learning scheme that learns to transfer appearance details efficiently for pose-guided synthesis. We further propose a unified flow-based network design for coarse-to-fine pose-guided synthesis. Experiments on the DeepFashion dataset and other datasets show that our approach robustly generates better texture and details in comparison with the existing state-of-the-art methods. Our approach can generalize to new pose and clothing styles, and it is effective in real-world scenes.

%\vspace{-0.3cm}
%Aiming for the challenge large-scale ($8\times$) super-resolution problem, we propose an end-to-end reference-based super resolution network named as CrossNet, where the input is a low-resolution (LR) image and a high-resolution (HR) reference image that shares similar view-point, the output is the super-resolved (4x or 8x) result of LR image. The pipeline of CrossNet is full-convolutional, containing encoder, cross-scale warping, and decoder respectively. Extensive experiment on several large-scale datasets demonstrate the superior performance of CrossNet (around 2dB-4dB) compared to previous methods. More importantly, CrossNet achieves a speedup of more than 100 times compared to existing RefSR approaches, allowing the model to be applicable for real-time applications.

\appendices
\section{Adaptation of FlowNetS}
\label{appendix:flownet}
To implement the flow estimator function $\Flow()$ from Eq.~\ref{equ:flow}, we use the FlowNetS network structure. However, several adaptations are made. First, we reduce the channel of each convolution/deconvolution layer to $64$ for memory efficiency. Second, to improve the flow definition at scale $0$, the $\times 4$ bilinear upsampling layer at the end of the original FlowNetS is replaced by two $\times 2$ U-Net upsampling modules.

\section{Code for gated multiplicative attention filtering}
\label{appendix:filtering}
We show that the gated multiplicative attention filtering 
\begin{equation*}
% \label{eq:multiplicative_filter}
\mathbf{f}_{s\rightarrow t}^{(l)\prime} = \mathbf{f}^{(l)}_{s\rightarrow t} \odot \sigma(\mathbf{f}^{(l)\top}_{s\rightarrow t} \mathbf{W}^{(l)} \mathbf{f}^{(l)}_t),
\end{equation*}
from Eq.~\ref{eq:multiplicative_filter} can be implemented using 3 lines of code in PyTorch:

\begin{algorithm}
\caption{Gated multiplicative attention filtering}
\begin{algorithmic}[1]
 \renewcommand{\algorithmicrequire}{\textbf{Input:}}
 \renewcommand{\algorithmicensure}{\textbf{Output:}}
 \REQUIRE $\mathbf{f}_{s\rightarrow t}^{(l)\prime}, \mathbf{f}^{(l)}_t$
 \ENSURE  $\mathbf{f}_{s\rightarrow t}^{(l)\prime}$
 \\ \textit{compute filter $\sigma(\mathbf{f}^{(l)\top}_{s\rightarrow t} \mathbf{W}^{(l)} \mathbf{f}^{(l)}_t)$} :
  \STATE \texttt{att = torch.sum(conv\_W($\mathbf{f}_{s\rightarrow t}^{(l)}$) *$\mathbf{f}^{(l)}_t$, 1)}
  \STATE \texttt{att = torch.sigmoid(att)}
   \\ \textit{perform filtering} :
   \STATE \texttt{$\mathbf{f}_{s\rightarrow t}^{(l)\prime}$ = torch.mul($\mathbf{f}_{s\rightarrow t}^{(l)}$, att)}\
 \RETURN $\mathbf{f}_{s\rightarrow t}^{(l)\prime}$ 
 \end{algorithmic} 
\end{algorithm}

where function \texttt{conv\_W()} defines a $1 \times 1$ convolutional operation with its trainable parameters $\mathbf{W}^{(l)}$.

% Can use something like this to put references on a page
% by themselves when using endfloat and the captionsoff option.
\ifCLASSOPTIONcaptionsoff
  \newpage
\fi

% trigger a \newpage just before the given reference
% number - used to balance the columns on the last page
% adjust value as needed - may need to be readjusted if
% the document is modified later
%\IEEEtriggeratref{8}
% The "triggered" command can be changed if desired:
%\IEEEtriggercmd{\enlargethispage{-5in}}

% references section

% can use a bibliography generated by BibTeX as a .bbl file
% BibTeX documentation can be easily obtained at:
% http://mirror.ctan.org/biblio/bibtex/contrib/doc/
% The IEEEtran BibTeX style support page is at:
% http://www.michaelshell.org/tex/ieeetran/bibtex/
%\bibliographystyle{IEEEtran}
% argument is your BibTeX string definitions and bibliography database(s)
%\bibliography{IEEEabrv,../bib/paper}
%
% <OR> manually copy in the resultant .bbl file
% set second argument of \begin to the number of references
% (used to reserve space for the reference number labels box)
\newpage
{
	\bibliographystyle{ieee}
	\bibliography{egbib}
}

%\begin{thebibliography}{1}
%\bibitem{IEEEhowto:kopka}
%H.~Kopka and P.~W. Daly, \emph{A Guide to \LaTeX}, 3rd~ed.\hskip 1em plus
%  0.5em minus 0.4em\relax Harlow, England: Addison-Wesley, 1999.
%
%\end{thebibliography}

% biography section
% 
% If you have an EPS/PDF photo (graphicx package needed) extra braces are
% needed around the contents of the optional argument to biography to prevent
% the LaTeX parser from getting confused when it sees the complicated
% \includegraphics command within an optional argument. (You could create
% your own custom macro containing the \includegraphics command to make things
% simpler here.)
%\begin{IEEEbiography}[{\includegraphics[width=1in,height=1.25in,clip,keepaspectratio]{mshell}}]{Michael Shell}
% or if you just want to reserve a space for a photo:

\vspace{10mm}
\begin{IEEEbiography}[{\includegraphics[width=1in,height=1.25in,clip,keepaspectratio]{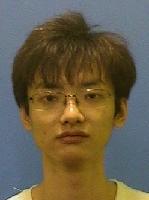}}]{Haitian Zheng}
Haitian Zheng received the B.Sc. and the M.Sc. degrees in electronics engineering and informatics science from the University of Science and Technology of China, under the supervision of Prof. Lu Fang, in 2012 and 2016, respectively. He is currently pursuing the PhD degree with the Computer Science Department, University of Rochester, under the supervision of Prof. Jiebo Luo.
His research interests include computer vision and machine learning.
\end{IEEEbiography}

% if you will not have a photo at all:
\begin{IEEEbiography}[{\includegraphics[width=1in,height=1.25in,clip,keepaspectratio]{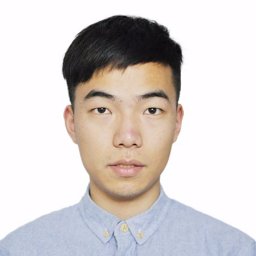}}]{Lele Chen}
Lele is a Ph.D candidate advised by Prof. Chenliang Xu in URCS. He received his M.S. degree in Computer Science from University of Rochester in 2018 and B.S. degree in Computer Science from Donghua University in 2016. His research interests are multimodal modeling and video object detection/segmentation.
\end{IEEEbiography}

\begin{IEEEbiography}[{\includegraphics[width=1in,height=1.25in,clip,keepaspectratio]{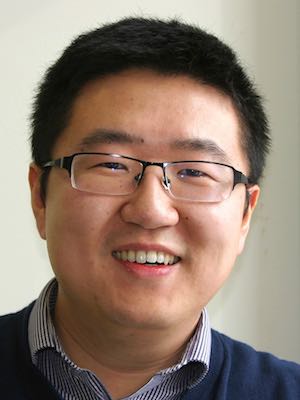}}]{Chenliang Xu}
Chenliang Xu is an Assistant Professor in the Department of Computer Science at the University of Rochester. He received his Ph.D. degree from the University of Michigan in 2016, the MS degree from SUNY Buffalo in 2012, both in Computer Science, and the BS degree in Information and Computing Science from Nanjing University of Aeronautics and Astronautics in 2010. He is the recipient of multiple NSF awards including BIGDATA 2017, CDS\&E 2018, and IIS Core 2018, the University of Rochester AR/VR Pilot Award 2017, Tencent Rhino-Bird Award 2018, the Best Paper Award at Sound and Music Computing 2017, and an Open Source Code Award in CVPR 2012. Xu has authored more than 30 peer-reviewed papers in venues such as IJCV, CVPR, ICCV, ECCV, IJCAI, and AAAI on topics of his research interest including computer vision and its relations to natural language, robotics, and data science. He co-organized the CVPR 2017 Workshop on video understanding and has served as a PC member and a regular reviewer for various international conferences and journals.
\end{IEEEbiography}

% insert where needed to balance the two columns on the last page with
% biographies
%\newpage

\begin{IEEEbiography}[{\includegraphics[width=1in,height=1.25in,clip,keepaspectratio]{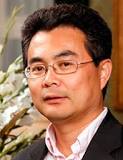}}]{Jiebo Luo}
Jiebo Luo (S93, M96, SM99, F09) joined the Department of Computer Science at the University of Rochester in 2011, after a prolific career of over 15 years with Kodak Research. He has authored over 400 technical papers and holds over 90 U.S. patents. His research interests include computer vision, machine learning, data mining, social media,and biomedical informatics. He has served as the Program Chair of the ACM Multimedia 2010, IEEE CVPR 2012, ACM ICMR 2016, and IEEE ICIP 2017, and on the Editorial Boards of the IEEE
TRANSACTIONS ON PATTERN ANALYSIS AND MACHINE INTELLIGENCE, IEEE TRANSACTIONS ON MULTIMEDIA, IEEE TRANSACTIONS ON CIRCUITS AND SYSTEMS FOR VIDEO TECHNOLOGY,  IEEE TRANSACTIONS ON BIG DATA,  Pattern Recognition, Machine Vision and Applications, and ACM Transactions on Intelligent Systems and Technology. He is also a Fellow of ACM, AAAI, SPIE and IAPR.
\end{IEEEbiography}

% You can push biographies down or up by placing
% a \vfill before or after them. The appropriate
% use of \vfill depends on what kind of text is
% on the last page and whether or not the columns
% are being equalized.

%\vfill

% Can be used to pull up biographies so that the bottom of the last one
% is flush with the other column.
%\enlargethispage{-5in}

% that's all folks
\end{document}